\documentclass[sigconf]{acmart}
\AtBeginDocument{%
  }

\copyrightyear{2026}
\acmYear{2026}
\setcopyright{cc}
\setcctype{by}
\acmConference[KDD 2026] {Proceedings of the 32nd ACM SIGKDD Conference on Knowledge Discovery and Data Mining V.2}{August 9--13, 2026}{Jeju Island, Republic of Korea.}
\acmBooktitle{Proceedings of the 32nd ACM SIGKDD Conference on Knowledge Discovery and Data Mining V.2 (KDD 2026), August 9--13, 2026, Jeju Island, Republic of Korea}
\acmISBN{979-8-4007-2259-2/2026/08}
\acmDOI{10.1145/3770855.3817729}
\usepackage{booktabs}
\usepackage{multirow}
\usepackage{makecell}
\usepackage{xspace}
\usepackage[ruled,vlined,linesnumbered]{algorithm2e}
\usepackage{graphicx}
\usepackage{subcaption}
\usepackage{pifont, bbding}
\usepackage{array}
\usepackage{amsmath}
\usepackage{hyperref}
\begin{document}

\title{Online Irregular Multivariate Time Series Forecasting via Uncertainty-Driven Dual-Expert Calibration}


\author{Haonan Wen}
\affiliation{  
  \institution{Key Laboratory of Big Data \&
Artificial Intelligence in Transportation (Beijing Jiaotong University), Ministry of Education}
  \institution{School of Computer Science and Technology, Beijing Jiaotong University}
  \city{Beijing}
  \country{China}
}
\email{hn.wen@bjtu.edu.cn}

\author{Hanyang Chen}
\affiliation{%
  \institution{Key Laboratory of Big Data \&
Artificial Intelligence in Transportation (Beijing Jiaotong University), Ministry of Education}
  \institution{School of Computer Science and Technology, Beijing Jiaotong University}
  \city{Beijing}
  \country{China}}
\email{hanyangchen@bjtu.edu.cn}

\author{Songhe Feng} 
\authornote{Corresponding author.}
\affiliation{%
  \institution{School of Computer Science and Technology, Beijing Jiaotong University}
  \city{Beijing}
  \country{China}}
\affiliation{%
  \institution{Tangshan Research Institute, Beijing Jiaotong University}
  \city{Tangshan}
  \country{China}}
\email{shfeng@bjtu.edu.cn}

\renewcommand{\shortauthors}{Haonan Wen, Hanyang Chen, \& Songhe Feng}
\newcommand{\undercal}{\textsc{Under-Cali}\xspace}

\newcommand{\rxmark}{\textcolor[RGB]{0,0,0}{\ding{55}}}
\newcommand{\gcmark}{\textcolor[RGB]{0,0,0}{\ding{51}}}
\begin{abstract}
  Irregular multivariate time series (IMTS) forecasting is critical in many real-world applications, where time series are irregularly sampled and exhibit dynamically evolving missingness patterns. Although existing methods perform well in offline settings, they often suffer from significant performance degradation when deployed online due to dynamic shifts in data distribution. Maintaining forecasting capability in such dynamic scenarios typically necessitates online adaptation techniques. Since irregular sampling fundamentally undermines temporal continuity and periodicity, we cannot leverage these widely studied characteristics from regular MTS for online learning. To this end, we study the problem of online IMTS forecasting and propose \undercal, an uncertainty-driven dual-expert calibration framework consisting of three core components: an uncertainty estimator, a dual-expert calibration module, and an adaptive routing module. We design an uncertainty estimator that serves as the core control signal to jointly manage inference and adaptation processes. In our framework, the uncertainty estimator first assesses uncertainty for each incoming batch. The adaptive routing module then directs samples with high uncertainty to the unreliable expert for calibration, while low uncertainty samples remain with the reliable expert. Subsequently, the system updates the reliable expert and the uncertainty estimator using well-calibrated reliable samples, and updates the unreliable expert with challenging samples, enabling stable and efficient online learning. \undercal keeps the source forecasting model frozen and performs adaptation only through a lightweight, model-agnostic calibration module, enabling efficient adaptation. Extensive experiments on IMTS benchmarks demonstrate consistent improvements with low computational cost. Our code is available at \url{https://github.com/HaonanWen/Under-Cali}.

\end{abstract}



\maketitle

\section{Introduction}
\label{sec:introduction}

\begin{figure}
    \centering
    \begin{subfigure}[b]{0.45\linewidth}
        \centering
        \includegraphics[width=\linewidth, clip]{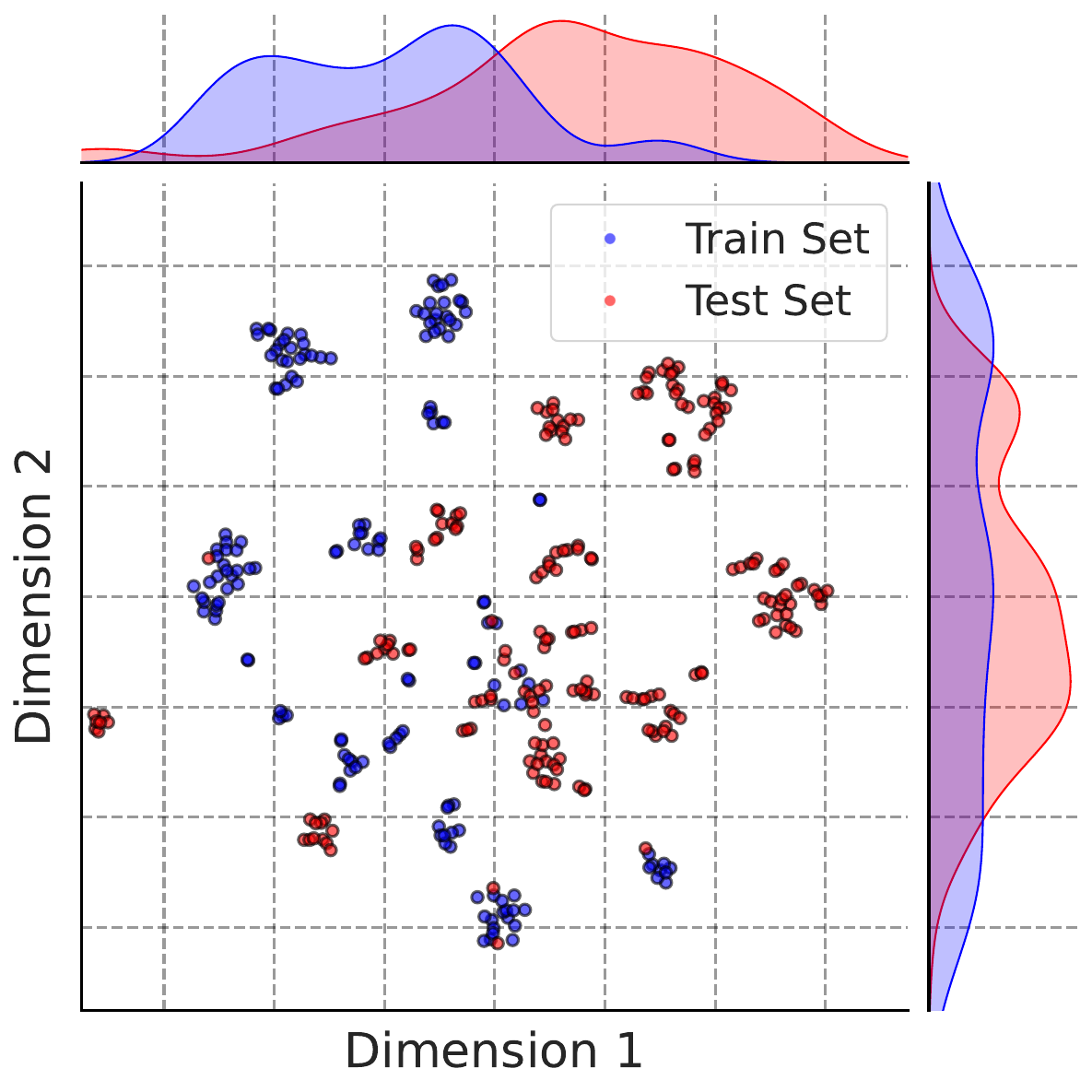}
        \caption{Human Activity}
        \label{fig:distribution_shift_human}
    \end{subfigure}
    \hfill
    \begin{subfigure}[b]{0.45\linewidth}
        \centering
        \includegraphics[width=\linewidth, clip]{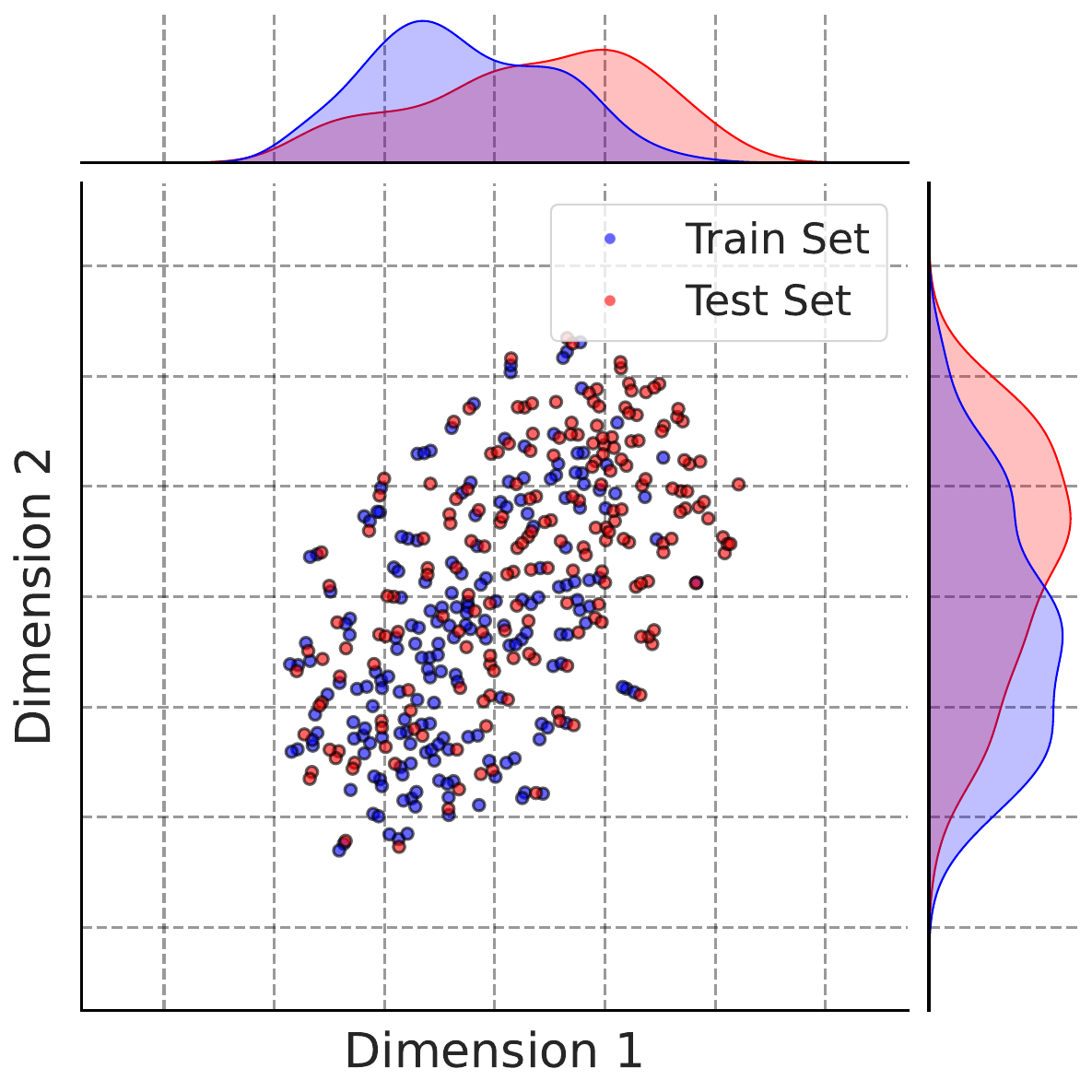}
        \caption{USHCN}
        \label{fig:distribution_shift_USHCN}
    \end{subfigure}

    \caption{Train–test distribution shifts in the Human Activity and USHCN datasets.}
    \label{fig:distribution_shift}
\end{figure}

Irregular multivariate time series (IMTS) forecasting is critical in applications such as healthcare, biomechanics, climate, astronomy and finance \cite{vio2013irregular, yao2018applying, shukla2020survey, zhang2024self}, where time series are captured at irregular sampling intervals and temporal misalignment. While substantial progress has been made in offline IMTS forecasting using continuous-time models, graph-based representations, and attention mechanisms \cite{rubanova2019latent, de2019gru, bilovs2021neural, schirmer2022modeling, zhang2024irregular, yalavarthi2024grafiti, zhang2023warpformer, li2025hyperimts}, the parameters in these methods will be frozen once training is completed, and thus they lack the ability to continuously adapt to evolving data. 

However, distribution shifts between the training and testing data often occur in many real-world application scenarios. It stems from the evolving data distribution and observation process caused by changes in subjects, devices, environments, or operational policies, leading to significant performance degradation after deployment. We present a data analysis in Figure \ref{fig:distribution_shift} to verify the presence of distribution shifts. While several online learning studies have been proposed to address this issue, most of them focus on regular time series \cite{lau2025fast,wen2023onenet}, where dense temporal continuity and periodicity provide reliable learning signals. These assumptions break down in IMTS, where temporal dependencies are sparse and inconsistent, observation intervals are highly variable, and missingness itself is informative and non-stationary \cite{zhang2024irregular}. As a result, directly applying existing online time series adaptation methods is not an appropriate solution. This gap motivates us to formally define and study the problem of online irregular multivariate time series forecasting, which models should continuously generate predictions in a sequential manner, achieve online adaptation under distribution shifts, and simultaneously handle complex issues such as irregular sampling and dynamically evolving missingness patterns.

In this paper, we directly focus on the uncertainty caused by distribution shifts, instead of the temporal continuity and periodicity. To better measure the forecasting uncertainty, we assume that a high forecasting uncertainty for a sample often indicates that the model has not learned or seen this underlying distribution, signaling a distribution shift. In such cases, online adaptation can be leveraged to update the model or the parameters of adapters, allowing it to assimilate the new data distribution and thereby achieve better performance on subsequent test data.

Building on the insight, we propose \undercal, an Uncertainty-driven Dual-Expert Calibration framework for online IMTS forecasting. Concretely, it consists of an Uncertainty Estimator (UE), a Dual-Expert Gated Distribution Calibrator (GDC), and an Adaptive Routing Module (ARM). We define uncertainty operationally as the normalized prediction error under the current data distribution, rather than a Bayesian confidence estimate. Intuitively, increased uncertainty reflects a mismatch between the current and the learned representation, providing a principled and reliable indicator for inference and adaptation. During inference, the UE first assesses uncertainty for each incoming batch per sample. Then, guided by their uncertainty scores, the samples are allocated to different gated distribution calibration experts for input and output calibration. The GDC separates them into two isolated pathways where a reliable expert refines inputs and predictions for familiar patterns with low uncertainty, and an unreliable expert cautiously adapts to novel or anomalous patterns with high uncertainty. This separation prevents severe out-of-distribution samples from corrupting stable forecasting patterns, addressing a core failure mode of existing online learning approaches. In addition, blindly adapting at every batch in the online adaptation may amplify gradient noise and lead to catastrophic interference, while an overly conservative adaptation strategy fails to effectively track distribution shifts. The ARM autonomously determines when to adapt the GDC and the UE. If adaptation is necessary, the system updates the reliable expert and the UE using well-calibrated reliable samples, and updates the unreliable expert with challenging samples. 

Importantly, predictions for each incoming batch are generated using parameters updated from previous batches, and the ground truth is used exclusively for adaptation, satisfying the online learning protocol. The entire framework is resource-efficient, updates only a small fraction of parameters, and is model-agnostic, seamlessly compatible with any existing IMTS forecaster. In summary, our contributions are as follows:
\begin{itemize}

\item We are the first to define the problem of Online Irregular Multivariate Time Series Forecasting and identify key challenges that distinguish it from existing settings.
\item We propose \undercal, an uncertainty-driven dual-expert calibration framework that enables principled and stable online adaptation for IMTS. We design a dual-expert gated distribution calibrator that performs differentiated calibration based on sample uncertainty scores quantified by the uncertainty estimator.
\item We introduce an adaptive routing module that governs the autonomously on-demand triggering and sample allocating to distinct calibration experts for stable online forecasting.
\item We conduct extensive experiments on IMTS benchmarks and architectures, demonstrating that \undercal consistently achieves superior performance compared to those without online adaptation.

\end{itemize}

\section{Related Work}
\label{sec:related-work}

\subsection{IMTS Forecasting}
Recent advancements in deep learning have significantly propelled the field of Irregular Multivariate Time Series (IMTS) forecasting. From the perspective of data representation for model input, existing approaches can be broadly classified into two paradigms: padding-based and non-padding-based methods.

Padding-based methods first convert IMTS into regular grid-like structures by inserting placeholders such as zeros, imputed values, or repeated observations along the time axis for each channel. This alignment facilitates the use of standard neural architectures. A prominent line of work leverages continuous-time models parameterized by Neural Ordinary Differential Equations (Neural ODEs) or their variants \cite{rubanova2019latent, bilovs2021neural, schirmer2022modeling}, which treat observations as points on a learned latent trajectory, inherently accommodating irregular intervals. Another branch adapts discrete-time architectures, such as Transformers \cite{zhang2023warpformer} and Graph Neural Networks \cite{zhang2021graph, luo2024knowledge, han2024bigst}, to the padded representations to capture temporal and cross-variable dependencies. While effective, these methods often incur increased computational cost due to processing filler data, and the padding operation itself may distort the original temporal dynamics and sparsity patterns.

In contrast, non-padding-based methods aim to model the raw, sparse IMTS directly using more flexible data structures. SeFT \cite{horn2020set} employs set representations where observations are treated as unordered elements, focusing on permutation-invariant aggregation. GraFITi \cite{yalavarthi2024grafiti} treats the series as a bipartite graph, connecting separate sets of variable nodes and time nodes. A recent advancement is the hypergraph-based approach, exemplified by HyperIMTS \cite{li2025hyperimts}, which models all observed values as nodes interconnected by temporal and variable hyperedges. This structure enables direct message passing among unaligned observations, offering a more unified and expressive representation. However, these approaches often face challenges in addressing the distribution shift issue after model deployment and lack the capability for online adaptation.

\subsection{Online Learning for Time Series}

Research on online time series forecasting has evolved from traditional statistical models toward more complex deep learning architectures. Early work \cite{anava2013online} updated ARMA model parameters online using regret minimization techniques, though such models struggle to capture complex temporal dependencies. Subsequently, researchers have proposed deep learning-based online models, such as OneNet \cite{wen2023onenet}, which integrates cross-variable and cross-time modeling and FSNet \cite{pham2022learning} introduces a dual-stream mechanism to leverage both immediate and delayed feedback. To address shifting data distributions, the Detect-and-Adapt method \cite{zhang2024addressing} was introduced to explicitly detect concept drift and adjust the model. Additionally, some studies have focused on reducing the computational cost of online updates. DSOF \cite{lau2025fast} employs an adapter module and experience replay to efficiently fine-tune the model with limited data. Proceed \cite{zhao2025proactive} proposes a proactive adaptation strategy that adjusts model weights using low-rank matrices. Concurrently, TAFAS \cite{kim2025battling} proposes a test-time adaptation framework that utilizes partially observed ground truths to proactively calibrate predictions.

These methods primarily target regularly sampled time series. However, existing work has not yet systematically studied the online forecasting scenario for IMTS, where irregular sampling, asynchronous variables, and dynamically evolving missingness fundamentally undermine temporal continuity and periodicity, making direct application of existing online adaptation techniques inefficient or unstable. This paper aims to fill this gap.

\section{Preliminary}

We consider an IMTS dataset $\mathcal{D} := \{\mathcal{O}_i|i = 1, ..., N\}$ consisting of $N$ samples, where $\mathcal{O}_i$ is the $i$-th sample. Each sample $\mathcal{O}_i$ with a total of $L$ timestamps and $C$ variables can be represented by triplet $\mathcal{O}_i = (\mathcal{T}_i,\mathcal{V}_i,\mathcal{M}_i)$. Here, $\mathcal{T}_i = \cup _{c=1}^C[l^c_j]_{j=1}^{L_c}\in \mathbb{R}^L$ denotes the chronological unique timestamps of all observations within $C$ variables. Matrix $\mathcal{V}_i =[[v_l^c]_{c=1}^C]_{l=1}^L \in \mathbb{R}^{L \times C}$ records the observed values or ‘NA’ if unobserved. Matrix $\mathcal{M}_i = [[m_l^c]_{c=1}^C]_{l=1}^L\in \mathbb{R}^{L \times C}$ represents a masking matrix, where $m_l^c$ is 1 if $v_l^c$ is observed, otherwise zero. 

For the forecasting task, given a split timestamp $l_h\in [1, L]$, an IMTS sample can be divided into a lookback window $\mathcal{X}_i := (\mathcal{T}_{i,x}, \mathcal{V}_{i,x}, \mathcal{M}_{i,x})$ and a forecast window $\mathcal{Y}_i := (\mathcal{T}_{i,y}, \mathcal{V}_{i,y}, \mathcal{M}_{i,y})$. The forecast query $\mathcal{Q}_i$ is the combination of $\mathcal{T}_{i,y}$ and $\mathcal{M}_{i,y}$ within the forecast window, \textit{i.e.}, $\mathcal{Q}_i := (\mathcal{T}_{i,y}, \mathcal{M}_{i,y})$. In this work, given the lookback window $\mathcal{X}_i$ and forecast query $\mathcal{Q}_i$, a source forecasting model $f_\theta(\cdot)$ can predict the corresponding observed values $\hat{\mathcal{V}}_{i,y}$:
\begin{equation}
f_\theta(\mathcal{X}_i,\mathcal{Q}_i) \rightarrow \hat{\mathcal{V}}_{i,y}. 
\end{equation}

\section{Methodology}

\begin{figure*}
    \centering
    \includegraphics[width=\linewidth]{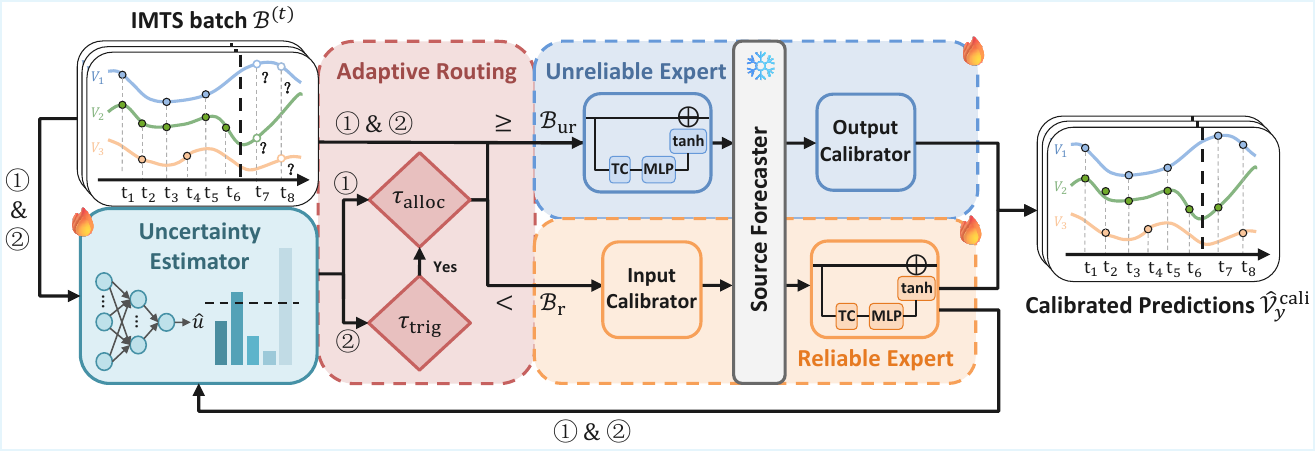}
    \caption{The framework of \undercal. For each incoming batch, (1) samples are calibrated by the reliable expert first for preliminary predictions, which are then fed into the uncertainty estimator along with input to compute per-sample uncertainty scores. Guided by these scores, the adaptive routing module directs high-uncertainty samples to the unreliable expert for secondary calibration to obtain final predictions, while low-uncertainty samples retain their preliminary calibrated results. (2) After prediction, if the adaptive routing module determines an adaptation is required, the reliable expert and uncertainty estimator are updated using low-uncertainty samples, while the unreliable expert is updated with challenging samples.}
\label{fig:undercal_framework}
\end{figure*}

As illustrated in Figure \ref{fig:undercal_framework}, \undercal operates in an online manner and is built upon three core components.
The uncertainty estimator, introduced in Section \ref{sec:uncertainty_estimator}, quantifies the forecasting uncertainty of each sample.
The dual-expert gated distribution calibrator, detailed in Section \ref{sec:gdc}, performs expert-isolated input and output calibration.
The adaptive routing module, described in Section \ref{sec:adaptive_routing}, governs when to trigger adaptation and how to allocate samples to the calibration experts.
In addition, the online adaptation process is introduced for online deployment in section \ref{sec:online_process}.

\subsection{Uncertainty Estimator}
\label{sec:uncertainty_estimator}

Due to the irregular nature of IMTS, the temporal continuity and periodicity commonly exploited in recent multivariate time series studies \cite{kim2025battling} to determine model update timing are no longer accessible. Therefore, we resort to uncertainty estimation as an alternative indicator to quantify the degree of distribution shift and determine when to update the model. 
However, conventional methods such as MC dropout \cite{gal2016dropout} depend on dropout layers and cannot generalize to arbitrary model architectures. To enable model-agnostic adaptation, we train an Uncertainty Estimator (UE) to estimate the uncertainty score, using it as a measure of uncertainty \cite{liu2025improving}. It is necessary to clarify that we define uncertainty operationally
as the normalized prediction error under the current data distribution, rather than a Bayesian confidence estimate. Importantly, UE is not used to directly refine predictions, but solely serves as a control signal for online adaptation decisions.

Specifically, we implement UE as a neural network $g_{\phi}$ that learns a mapping from the joint representation of input time series and the prediction results to the uncertainty score, which can be written as follows,
\begin{equation}
    \hat{u}_i = g _\phi (\mathcal{X}_i, \hat{\mathcal{V}}_{i,y}),
    \label{eq:ue-prediction}
\end{equation}
where $\mathcal{X}_i$ is the flattened lookback window, $\hat{\mathcal{V}}_{i,y}$ is the the model prediction, and $\hat{u}_i \in [0,1]$ is the estimated uncertainty score. The ground truth of the uncertainty score $u_i$ is computed as the normalized Mean Squared Error (MSE) between the predicted values $\hat{\mathcal{V}}_{i,y}$ and observed values $\mathcal{V}_{i,y}$, which can be formulated as follows,
\begin{equation}
u_i = \frac{\Delta - \Delta_\mathrm{min}}{\Delta_\mathrm{max}-\Delta_\mathrm{min}}, \Delta = \parallel(\hat{\mathcal{V}}_{i,y} - \mathcal{V}_{i,y}) \odot \mathcal{M}_{i,y} \parallel^2_2,
\label{eq:ue-target}
\end{equation}
where $\Delta_\mathrm{min}$, $\Delta_\mathrm{max}$ are the minimum value and maximum value in running statistics. Notably, the observation mask $\mathcal{M}_{i,y}$ allows UE to learn the missingness pattern of the target series and distinguish whether potential error stems from poor forecasting or inherent sparsity of the ground truth. 

The UE is pre-trained offline first to regress the normalized MSE value of the source forecaster with the Mean Absolute Error (MAE) loss function, which ensures that the estimated uncertainty aligns with the actual forecasting error. 
This direct supervision ensures that a higher output value $\hat{u}_i$ correlates directly with a higher expected prediction error and, by proxy, higher estimated forecasting uncertainty, signaling a significant distribution shift.

During online adaptation, both the ground-truth and estimated uncertainty scores are computed using the calibrated prediction $\hat{\mathcal{V}}_{i,y}^{\mathrm{cali}}$ produced by the reliable expert introduced in Section \ref{sec:gdc}, rather than by the source forecaster.
\begin{equation}
    \hat{u}_i = g _\phi (\mathcal{X}_i, \hat{\mathcal{V}}_{i,y}^\mathrm{cali}).
    \label{eq:ue-prediction-online}
\end{equation}
\begin{equation}
u_i = \frac{\Delta - \Delta_\mathrm{min}}{\Delta_\mathrm{max}-\Delta_\mathrm{min}}, \Delta = \parallel(\hat{\mathcal{V}}_{i,y}^\mathrm{cali} - \mathcal{V}_{i,y}) \odot \mathcal{M}_{i,y} \parallel^2_2,
\label{eq:ue-target-online}
\end{equation}
After the ground truth becomes available, the UE will be updated online to improve uncertainty estimation for subsequent batches, ensuring fair and stable evaluation under distribution shift. We detail the online adaptation process in Section \ref{sec:online_process}.

\subsection{Dual-Expert Gated Distribution Calibrator}
\label{sec:gdc}

Rather than fine-tuning the source forecaster directly, we choose to introduce a calibration module to achieve more efficient adaptation with lower resource consumption. However, updating the calibrator with high-uncertainty samples may lead to unstable adaptation on subsequent low-uncertainty samples. 

To this end, we introduce a model-agnostic Dual-Expert Gated Distribution Calibrator (GDC), which comprises two experts with separate parameters, \textit{i.e.}, a reliable expert $e_{\psi_\mathrm{r}}$ and an unreliable expert $e_{\psi_\mathrm{ur}}$. The reliable expert $e_{\psi_\mathrm{r}}$ calibrates low-uncertainty samples, which are close to the trained distribution and where the forecaster already performs well, applying fine-grained adjustments. In contrast, the unreliable expert $e_{\psi_\mathrm{ur}}$ calibrates high-uncertainty samples, which may be anomalies or reflect a significant distribution shift, requiring cautious adaptation to prevent overfitting. The two experts share an identical structure but serve distinct functions. They play complementary roles in handling samples of different uncertainty. By isolating the calibration pathways, \undercal ensures that learning from challenging samples with high uncertainty scores does not corrupt the precise calibration maintained for reliable patterns of low uncertainty samples. Besides, each expert can be tuned with tailored learning rates, allowing the system to learn efficiently from both high-quality and noisy data. For brevity, the index subscript $i$ is omitted in this section.

To concretize the design, each expert $e_{\psi_j}, j \in \{\mathrm{r}, \mathrm{ur}\}$ consists of an input calibrator $e_{\psi_{j, \mathrm{in}}}$ at the front-end of the source forecaster $f_\theta$ and an output calibrator $e_{\psi_{j, \mathrm{out}}}$ at its back-end.
Based on the uncertainty score described in Section \ref{sec:uncertainty_estimator} and the adaptive routing module introduced in Section \ref{sec:adaptive_routing}, each sample is routed to the appropriate expert for targeted adjustment of both its input features and final predictions. In addition, each calibrator is built upon a variable-wise, gated temporal calibration model. Specifically, for an input tensor $\mathcal{V} \in \mathbb{R}^{L \times C}$ representing a window of $L$ timesteps and $C$ variables, the calibrator first applies a temporal calibration via learnable weights $\mathbf{W}_c \in \mathbb{R}^{L \times L}$ and biases $\mathbf{b}_c \in \mathbb{R}^{L} $ per variable to capture local temporal shifts. This is followed by a non-linear enhancement via an MLP to model complex distribution changes. A learnable, variable-wise gating vector $\upsilon \in \mathbb{R}^{C}$ dynamically modulates the strength of the calibration, which is finally applied in a residual manner to ensure training stability:
\begin{equation}
    e_{\psi_{j,k}}(\mathcal{V}) = \mathcal{V} + \mathrm{Tile}(\mathrm{tanh}(\upsilon)) \odot  \mathrm{MLP}(\mathrm{Concat}(\{\mathbf{W}_c\mathcal{V}_{:,c} + \mathbf{b}_c\}_{c=1}^C)),
    \label{eq:calibrator}
\end{equation}
where $k\in\{\mathrm{in},\mathrm{out}\}$, $\mathrm{Concat}$ denotes concatenation along the variable dimension, $\mathrm{Tile}(\cdot): \mathbb{R}^C \rightarrow \mathbb{R}^{L\times C}$ broadcasts the gating vector along the temporal dimension, and $\odot$ is element-wise multiplication. 

Combining with both the input calibrator and output calibrator, each expert $e_{\psi_j} $ applies calibrators in two stages. The input calibrator $e_{\psi_{j,\mathrm{in}}}$ transforms the raw input observation $\mathcal{V}_x$ into a calibrated input  $\mathcal{V}^\text{cali}_x$ that aligns better with the forecaster’s original training domain. After the source forecaster $f_\theta$ processes the calibrated input to produce a preliminary prediction $\hat{\mathcal{V}}_y$, the output calibrator $e_{\psi_{j,\mathrm{out}}}$ further adjusts this output to $\hat{\mathcal{V}}^\mathrm{cali}_{j,y}$ align with the evolving distribution. It can be formulated as follows,
\begin{equation}
    \hat{\mathcal{V}}_{j, y}^\mathrm{cali} = e_{\psi_{j,\mathrm{out}}}(f_\theta((\mathcal{T}_{x}, e_{\psi_{j,\mathrm{in}}}(\mathcal{V}_{x}), \mathcal{M}_{x}), \mathcal{Q})).
    \label{eq:GDC-form}
\end{equation}

The parameters of both experts are updated online, guided by the uncertainty scores and the adaptive routing module. To ensure stability, all learnable parameters in both experts include linear weights, biases, and the final layer weights of the MLP are initialized to zero, while the gating vector is initialized to small values near zero, \textit{e.g.}, 0.01, making each GDC initially approximate an identity mapping. This design preserves the source forecaster’s original knowledge until distribution shifts compel the respective expert to apply meaningful calibration.

\subsection{Adaptive Routing Module}
\label{sec:adaptive_routing}

To achieve adaptive routing and updating for samples with various uncertainty scores and automatic model updating, we propose the Adaptive Routing Module (ARM), leveraging uncertainty scores as the primary decision signal. The module coordinates the online adaptation process through two independent adaptive thresholds, where the allocation threshold $\tau_\mathrm{alloc}$ routes each sample to the appropriate calibration expert, and the trigger threshold $\tau_\mathrm{trig}$ determines whether adaptation is necessary. The index subscript $i$ is omitted for brevity in this section.

To support the distinct functions of the thresholds, we maintain two independent statistics sets, $s_\mathrm{alloc}=\{\mu_\mathrm{alloc}, \sigma_\mathrm{alloc}^2\}$ and $s_\mathrm{trig}=\{\mu_\mathrm{trig}, \sigma_\mathrm{trig}^2\}$, which respectively track the distributions of uncertainty scores. Each set is updated using an Exponential Moving Average (EMA), with smoothing factors 
$\alpha_\mathrm{alloc}$ and $\alpha_\mathrm{trig}$ for sets $s_\mathrm{alloc}$ and $s_\mathrm{trig}$, respectively. Specifically, the mean and variance in each set are updated iteratively per batch as follows,

\begin{equation}
\mu_j^{(t)} = 
\begin{cases} 
\bar{u}^{(t)}, & t = 1 \\ 
(1-\alpha_j) \cdot \mu_j^{(t-1)} + \alpha_j \cdot \bar{u}^{(t)}, & t > 1 
\end{cases},
\label{eq:statistics-mu}
\end{equation}

\begin{equation}
(\sigma^2_j)^{(t)} = 
\begin{cases} 
\tilde{u}^{(t)}, & t = 1 \\ 
(1-\alpha_j) \cdot (\sigma^2_j)^{(t-1)} + \alpha_j \cdot \tilde{u}^{(t)}, & t > 1 
\end{cases},
\label{eq:statistics-sigma}
\end{equation}
where $\bar{u}^{(t)}$ and $\tilde{u}^{(t)}$ represent the mean value and variance of the estimated uncertainty scores in the $t$-th batch, and $j \in \{\mathrm{alloc}, \mathrm{trig}\}$ indicates the corresponding threshold in this section.

With the statistics sets established, we further introduce the allocation and trigger thresholds. Specifically, the allocation threshold $\tau_\mathrm{alloc}$ serves as a dispatcher that routes each sample to the appropriate expert. For the incoming batch $t$, we first update the statistics set $s_\mathrm{alloc}$ using the uncertainty scores of the current batch, then compute the threshold $\tau_\mathrm{alloc}^{(t)}$ based on these updated statistics:
\begin{equation}
    \tau_\mathrm{alloc}^{(t)} = \mu^{(t)}_\mathrm{alloc} + \kappa_\mathrm{alloc} \cdot \sigma^{(t)}_\mathrm{alloc}, 
    \label{eq:alloc-update}
\end{equation}
where $\kappa_\mathrm{alloc}$ is a coefficient controlling the allocation sensitivity. We utilize the statistics from the $t$-th batch to enable timely feedback on threshold adjustment, thus facilitating more accurate allocation for the current batch. During inference, each sample is routed to either the reliable or unreliable expert by comparing its estimated uncertainty score $\hat{u}$ against the allocation threshold $\tau_\mathrm{alloc}^{(t)}$. 
Samples with $\hat{u} < \tau_\mathrm{alloc}^{(t)}$ are deemed reliable and calibrated by the reliable expert, while others are calibrated by the unreliable expert:
\begin{equation}
\hat{\mathcal{V}}^{\mathrm{cali}}_y =
\begin{cases}
e_{\psi_{\mathrm{r,out}}}(f_\theta((\mathcal{T}_{x}, e_{\psi_{\mathrm{r,in}}}(\mathcal{V}_{x}), \mathcal{M}_{x}), \mathcal{Q})), 
& \hat{u} < \tau_{\mathrm{alloc}}^{(t)}, \\[4pt]
e_{\psi_{\mathrm{ur,out}}}(f_\theta((\mathcal{T}_{x}, e_{\psi_{\mathrm{ur,in}}}(\mathcal{V}_{x}), \mathcal{M}_{x}), \mathcal{Q})), 
& \hat{u} \ge \tau_{\mathrm{alloc}}^{(t)}.
\end{cases}
\label{eq:route-samples}
\end{equation}
The calibrated outputs are the final predictions for this batch. Then the adaptation process performs the same routing strategy. This strategy enables specialized learning through precise sample routing and enables stable online adaptation.

Meanwhile, the trigger threshold $\tau_\mathrm{trig}$ acts as a sentinel monitoring environmental changes and determining when to initiate an update, thereby achieving sparse triggering. For a new batch $t$, an update is triggered only when the mean uncertainty of this batch $\bar{u}^{(t)}$ exceeds $\tau_\mathrm{trig}^{(t)}$. The threshold is computed as follows,
\begin{equation}
    \tau_\mathrm{trig}^{(t)} = \mu^{(t-1)}_\mathrm{trig} + \kappa_\mathrm{trig} \cdot \sigma^{(t-1)}_\mathrm{trig}, 
    \label{eq:trig-update}
\end{equation}
where $\kappa_\mathrm{trig}$ is a coefficient controlling the triggering sensitivity. If an update is judged necessary, \undercal then use the batch to update the GDC and UE. 

The ARM forms an intelligent decision core that efficiently allocates computational resources and learning signals. Furthermore, using EMA for thresholds allows the system to adaptively track the evolving uncertainty distribution online, eliminating the need for preset thresholds.

\subsection{Online Adaptation Process}
\label{sec:online_process}

\begin{algorithm}[t]
\caption{Online Adaptation of \undercal}
\label{alg:undercal_online}
\KwIn{
Source forecaster $f_{\theta}$,
uncertainty estimator $g_{\phi}$,
reliable expert $e_{\psi_\mathrm{r}}$,
unreliable expert $e_{\psi_\mathrm{ur}}$,
dataset $\mathcal{D}$.
}

\KwOut{Calibrated prediction $\hat{\mathcal{V}}_y^{\mathrm{cali}}$}

\For{each incoming batch $\mathcal{B}^{(t)} = \{(\mathcal{X}^{(t)}, \mathcal{Q}^{(t)})|t = 1,...,T\}$}{

\For{$t = 1$ to $T$}{

\textcolor{gray}{// Inference Stage}

Calibrate predictions with the expert $e_{\psi_\mathrm{r}}$ via Eq.\ref{eq:GDC-form}

Estimate uncertainty scores via Eq.\ref{eq:ue-prediction-online}

Update the allocation statistics set and allocation threshold via Eq.\ref{eq:statistics-mu}, \ref{eq:statistics-sigma} and \ref{eq:alloc-update} 

Route samples and recalibrate predictions via Eq.\ref{eq:route-samples}

\textcolor{gray}{// Adaptation Stage}

Update the trigger threshold via Eq.\ref{eq:trig-update}

\If{$\bar{u}^{(t)} > \tau_{\mathrm{trig}}^{(t)}$}{

Update reliable expert $e_{\psi_\mathrm{r}}$ via Eq.\ref{eq:expert-loss}

Update unreliable expert $e_{\psi_\mathrm{ur}}$ via Eq.\ref{eq:expert-loss}

Update uncertainty estimator $g_\phi$ via Eq.\ref{eq:ue-loss}

}

Update the trigger statistics set via Eq.\ref{eq:statistics-mu} and \ref{eq:statistics-sigma} 

}

}
\end{algorithm}

In this section, we further introduce the online adaptation process of our method.
For online deployment, \undercal processes each incoming batch into two separated stages: (1) an inference stage generating the calibrated predictions, and (2) an adaptation stage where parameters are updated for future batches. In the following, we provide a detailed description of these two stages.

First, in the inference stage, samples from each incoming batch are initially treated as reliable and processed by the reliable expert in GDC to obtain preliminary calibrated predictions. These predictions are then fed into the UE along with the input series to compute per-sample uncertainty scores $\hat{u}_i$. Guided by these uncertainty scores, the ARM directs high-uncertainty samples to the unreliable expert for secondary calibration, producing their final predictions, while low-uncertainty samples retain their preliminary calibrated results. The performance of the current batch is evaluated based on these final predictions.

Then, in the adaptation stage, we use the mean value of the batch-wise uncertainty scores $\bar{u}_i$ to compare with the trigger threshold $\tau_\mathrm{trig}$ to judge whether to adapt. If so, the allocation threshold completes sample routing, and then the parameters of the corresponding expert in GDC are updated. Specifically, the subset of samples in a batch, $\mathcal{B}_j$, is assigned to the corresponding expert $e_{\psi_j}$, where $j \in \{\mathrm{r}, \mathrm{ur}\}$ denotes the reliable and unreliable experts, respectively.
Then, we perform a few gradient steps to minimize the prediction error on the assigned samples with the loss function as follows,
\begin{equation} 
\mathcal{L}_{e}(\psi_j)= \frac{1}{|\mathcal{B}_j|} \sum_{(\mathcal{X}_i, \mathcal{Y}_i) \in \mathcal{B}_j} \parallel(\hat{\mathcal{V}}^\mathrm{cali}_{i,j,y} - \mathcal{V}_{i,y}) \odot \mathcal{M}_{i,y}\parallel ^2_2,
\label{eq:expert-loss}
\end{equation}
where $\hat{\mathcal{V}}^\mathrm{cali}_{i,j,y}$ is the calibrated prediction from the output calibrator of expert $e_{\psi_{j,\mathrm{out}}}$ via Equation \ref{eq:GDC-form}.
After updating the experts, we move to fine-tune the parameters $\phi$ of UE. To maintain stable uncertainty estimation, we only use the reliable samples with uncertainty scores below the allocation threshold $\tau_\mathrm{alloc}$, \textit{i.e.}, $\mathcal{B}_\mathrm{r}$, to update it. We then compute the loss between the uncertainty scores estimated by UE and the actual prediction errors of the calibrated predictions from the reliable expert:
\begin{equation}
\mathcal{L}_{g}(\phi) = \frac{1}{|\mathcal{B}_\mathrm{r}|} \sum_{(\mathcal{X}_i, \mathcal{Y}_i) \in \mathcal{B}_\mathrm{r}} \| \hat{u}_i - u_i \|_1,
\label{eq:ue-loss}
\end{equation}
where $\hat{u}_i$ is the uncertainty scores estimated by uncertainty estimator $g_\phi$ at Equation \ref{eq:ue-prediction-online}, and $u_i$ is computed from the calibrated prediction $\hat{\mathcal{V}}_{i,\mathrm{r},y}^{\mathrm{cali}}$ and the ground truth $\mathcal{V}_{i,y}$ via Equation \ref{eq:ue-target-online}. Restricting UE updates to reliable samples prevents feedback loops caused by noisy or anomalous predictions.

In summary, \undercal enables principled online adaptation for IMTS by combining uncertainty-guided adaptive routing with expert-isolated calibration. Both the experts in GDC and the UE are updated online sequentially, creating a synergistic adaptation loop where improved predictions lead to better uncertainty estimation, which in turn guides more effective updates. Without modifying the source forecaster, \undercal provides a robust and model-agnostic solution to online IMTS forecasting. The complete online adaptation procedure is summarized in Algorithm~\ref{alg:undercal_online}.

\section{Experiments}
\label{sec:experiments}

In this section, we conduct extensive experiments to evaluate the performance of our method. We first introduce the experimental setup in Section \ref{sec:experimental-setup}. Then, we present the overall performance in Section \ref{sec:main-results}. Moreover, we conduct an ablation study to analyze the effectiveness of each component in Section \ref{sec:ablation-study}. The parameter study is provided in Section \ref{sec:parameter-study} to the sensitivity of hyperparameters. Finally, the lookback length and horizon study, comparison with existing online forecasting and test-time adaptation methods, and case study are provided in Section \ref{sec:varying-lookback}, \ref{sec:comparison-with-Online} and \ref{sec:case-study}, respectively.

\subsection{Experimental Setup}
\label{sec:experimental-setup}

\paragraph{{Datasets.}} We conduct comprehensive experiments on the PyOmniTS benchmark \cite{li2025hyperimts}, including four publicly available benchmark datasets spanning diverse domains to evaluate our proposed method. 
Specifically, for the clinical domain, we use MIMIC \cite{johnson2016mimic} and PhysioNet \cite{silva2012predicting}, both collected from ICU patients during the first 48 hours of admission. 
The measurements in MIMIC are rounded to 30-minute intervals, while those in PhysioNet are rounded to 1-hour intervals. A Human Activity dataset containing millisecond-resolution biomechanics signals describing 3D positional variables and a 4-year subset (1996–2000) of the USHCN \cite{menne2016long} climate repository are also adopted in the experiments. We follow standard preprocessing procedures from prior work \cite{zhang2024irregular} for all four datasets. For all experiments, we adopt a consistent split ratio of 20\% for training, 5\% for validation, and 75\% for online learning. The training and validation sets are shuffled, while the test set for online learning is preserved in its original order to simulate a realistic scenario. 

\paragraph{{Baselines.}} To conduct comprehensive comparisons, twenty-one source forecasters are considered in the experiments, covering SOTA methods from (1) Multivariate time series forecasting: FEDformer \cite{zhou2022fedformer}, FreTS \cite{yi2023frequency}, iTransformer \cite{liu2023itransformer}, Ada-MSHyper \cite{shang2024ada}, PatchTST \cite{nie2022time}, Reformer \cite{kitaev2020reformer}, Informer \cite{zhou2021informer}, Crossformer \cite{zhang2023crossformer}, (2) Traffic forecasting: BigST \cite{han2024bigst}, (3) IMTS classification, imputation, and forecasting: PrimeNet \cite{chowdhury2023primenet}, SeFT \cite{horn2020set}, mTAN \cite{shukla2021multi}, NeuralFlows \cite{bilovs2021neural}, CRU \cite{schirmer2022modeling}, GNeuralFlow \cite{mercatali2024graph}, GRU-D \cite{che2018recurrent}, Warpformer \cite{zhang2023warpformer}, tPatchGNN \cite{zhang2024irregular}, GraFITi \cite{yalavarthi2024grafiti}, Hi-Patch \cite{luohi}, HyperIMTS \cite{li2025hyperimts}, (4) online forecasting and test-time adaptation: OneNet \cite{wen2023onenet}, FSNet \cite{pham2022learning}, D3A \cite{zhang2024addressing}, TAFAS \cite{kim2025battling}.

\paragraph{{Implementation details.}} All experiments are conducted on a Linux server equipped with NVIDIA GeForce RTX 3090 GPUs using PyTorch 2.0.1 with CUDA 11.7. All source forecasters are trained offline using the Adam optimizer for up to 300 epochs with early stopping (patience = 5) based on validation performance. The UE is trained offline to regress sample-level prediction errors of the source model. UE is optimized with Adam and an L1 loss for up to 300 epochs with early stopping (patience = 10). During the online process, uncertainty scores are tracked using exponential moving averages of the mean and variance, with the smoothing factor $\alpha_\mathrm{alloc} = 0.75$ and scaling coefficient $\kappa_\mathrm{alloc} = 0.25$ for sample allocation. Update triggering is controlled independently using EMA-based statistics with $\alpha_\mathrm{trig} = 0.25$ and $\kappa_\mathrm{trig} = 0.75$. When an update is triggered, each expert is updated using 5 inner optimization steps.

\subsection{Main Results}
\label{sec:main-results}

\begin{table*}
\centering
\caption{Test MSE (mean $\pm$ std) on irregular multivariate time series forecasting datasets with and without \undercal across various architectures under five seeds. Lower MSE indicates better performance. Best results are highlighted in bold.}
\label{tab:main_results_mse}
\resizebox{\linewidth}{!}{
\begin{tabular}{ccccccccc}
\toprule
Models & \multicolumn{2}{c}{MIMIC} & \multicolumn{2}{c}{PhysioNet} & \multicolumn{2}{c}{Human Activity} & \multicolumn{2}{c}{USHCN}\\
\cmidrule(lr){2-3} \cmidrule(lr){4-5} \cmidrule(lr){6-7} \cmidrule(lr){8-9}
+\undercal & \rxmark & \gcmark & \rxmark & \gcmark & \rxmark & \gcmark & \rxmark & \gcmark\\
\midrule
FEDformer & 0.7234$\pm$0.0036 & \textbf{0.6244$\pm$0.0045} & 0.4097$\pm$0.0013 & \textbf{0.3789$\pm$0.0008} & 0.4212$\pm$0.0176 & \textbf{0.3966$\pm$0.0184} & 0.4965$\pm$0.0339 & \textbf{0.4391$\pm$0.0093} \\
FreTS & 0.6361$\pm$0.0017 & \textbf{0.5998$\pm$0.0020} & 0.3904$\pm$0.0019 & \textbf{0.3690$\pm$0.0028} & 0.1553$\pm$0.0025 & \textbf{0.1494$\pm$0.0019} & 0.5950$\pm$0.0452 & \textbf{0.5784$\pm$0.0538} \\
BigST & 0.5758$\pm$0.0043 & \textbf{0.5679$\pm$0.0032} & 0.3701$\pm$0.0048 & \textbf{0.3535$\pm$0.0028} & 0.4118$\pm$0.0240 & \textbf{0.2999$\pm$0.0164} & 0.4474$\pm$0.0168 & \textbf{0.4447$\pm$0.0166} \\
iTransformer & 0.5979$\pm$0.0026 & \textbf{0.5816$\pm$0.0090} & 0.3929$\pm$0.0028 & \textbf{0.3720$\pm$0.0019} & 0.1417$\pm$0.0018 & \textbf{0.1335$\pm$0.0012} & 0.4363$\pm$0.0331 & \textbf{0.4164$\pm$0.0195} \\
Ada-MSHyper & 0.6542$\pm$0.0140 & \textbf{0.6331$\pm$0.0091} & 0.3914$\pm$0.0007 & \textbf{0.3743$\pm$0.0071} & 0.3647$\pm$0.0154 & \textbf{0.3002$\pm$0.0284} & 0.4953$\pm$0.0458 & \textbf{0.4476$\pm$0.0257} \\
PatchTST & 0.6147$\pm$0.0024 & \textbf{0.5762$\pm$0.0102} & \textbf{0.3748$\pm$0.0014} & 0.3953$\pm$0.0243 & 0.1294$\pm$0.0239 & \textbf{0.1249$\pm$0.0041} & 0.6334$\pm$0.0198 & \textbf{0.6039$\pm$0.0208} \\
Reformer & \textbf{0.7738$\pm$0.0031} & 0.9490$\pm$0.1902 & \textbf{0.4113$\pm$0.0009} & 0.4169$\pm$0.0083 & 1.0842$\pm$0.0834 & \textbf{0.9269$\pm$0.0344} & 0.4382$\pm$0.0043 & \textbf{0.4346$\pm$0.0037} \\
Informer & \textbf{0.7328$\pm$0.0064} & 0.8077$\pm$0.0459 & 0.4944$\pm$0.0026 & \textbf{0.4860$\pm$0.0083} & 0.5202$\pm$0.0336 & \textbf{0.4386$\pm$0.0264} & \textbf{0.4584$\pm$0.0047} & 0.4922$\pm$0.0120 \\
Crossformer & 0.5294$\pm$0.0039 & \textbf{0.5249$\pm$0.0026} & 0.3404$\pm$0.0018 & \textbf{0.3391$\pm$0.0013} & 0.2716$\pm$0.0237 & \textbf{0.2391$\pm$0.0224} & 0.5591$\pm$0.0076 & \textbf{0.5153$\pm$0.0150} \\
\midrule
PrimeNet & 0.9103$\pm$0.0001 & \textbf{0.8960$\pm$0.0023} & 0.8095$\pm$0.0001 & \textbf{0.7879$\pm$0.0028} & 4.0244$\pm$0.0060 & \textbf{3.5619$\pm$0.0512} & 0.9441$\pm$0.0045 & \textbf{0.8098$\pm$0.0094} \\
SeFT & 0.9035$\pm$0.0004 & \textbf{0.8821$\pm$0.0070} & 0.8057$\pm$0.0033 & \textbf{0.7577$\pm$0.0076} & 1.8232$\pm$0.0703 & \textbf{1.4135$\pm$0.2791} & 0.7708$\pm$0.0038 & \textbf{0.7113$\pm$0.0103} \\
mTAN & 1.2596$\pm$0.1443 & \textbf{0.8809$\pm$0.0415} & 0.4305$\pm$0.0071 & \textbf{0.4193$\pm$0.0061} & 0.3550$\pm$0.0189 & \textbf{0.3354$\pm$0.0209} & 0.5218$\pm$0.0335 & \textbf{0.4792$\pm$0.0179} \\
NeuralFlows & 0.8332$\pm$0.0133 & \textbf{0.8000$\pm$0.0229} & 0.4456$\pm$0.0064 & \textbf{0.4408$\pm$0.0066} & 0.4732$\pm$0.0464 & \textbf{0.4390$\pm$0.0239} & 0.5627$\pm$0.0100 & \textbf{0.5445$\pm$0.0131} \\
GNeuralFlow & 0.9046$\pm$0.0067 & \textbf{0.8814$\pm$0.0024} & 0.7493$\pm$0.0494 & \textbf{0.6647$\pm$0.0197} & 0.4225$\pm$0.0188 & \textbf{0.4070$\pm$0.0167} & 0.5860$\pm$0.0201 & \textbf{0.5750$\pm$0.0245} \\
CRU & 0.8097$\pm$0.0037 & \textbf{0.7870$\pm$0.0126} & 0.6649$\pm$0.0031 & \textbf{0.6253$\pm$0.0049} & 0.4070$\pm$0.0598 & \textbf{0.3488$\pm$0.0305} & 0.5771$\pm$0.0161 & \textbf{0.5685$\pm$0.0170} \\
GRU-D & 0.6634$\pm$0.0110 & \textbf{0.6510$\pm$0.0112} & 0.3759$\pm$0.0007 & \textbf{0.3742$\pm$0.0006} & 0.3735$\pm$0.0011 & \textbf{0.3680$\pm$0.0034} & 0.5215$\pm$0.0381 & \textbf{0.5070$\pm$0.0327} \\
Warpformer & 0.4722$\pm$0.0070 & \textbf{0.4683$\pm$0.0044} & 0.3040$\pm$0.0026 & \textbf{0.3028$\pm$0.0014} & 0.3942$\pm$0.0441 & \textbf{0.3081$\pm$0.0346} & 0.4338$\pm$0.0081 & \textbf{0.4295$\pm$0.0085} \\
tPatchGNN & 0.4943$\pm$0.0128 & \textbf{0.4872$\pm$0.0099} & 0.3080$\pm$0.0007 & \textbf{0.3066$\pm$0.0008} & 0.1378$\pm$0.0873 & \textbf{0.1262$\pm$0.0732} & 0.4589$\pm$0.0063 & \textbf{0.4538$\pm$0.0092} \\
Hi-Patch & 0.5073$\pm$0.0088 & \textbf{0.5039$\pm$0.0063} & 0.3433$\pm$0.0021 & \textbf{0.3405$\pm$0.0017} & 1.1308$\pm$0.6606 & \textbf{1.0956$\pm$0.6544} & 0.4343$\pm$0.0137 & \textbf{0.4330$\pm$0.0138} \\
GraFITi & 0.4531$\pm$0.0030 & \textbf{0.4482$\pm$0.0040} & 0.2996$\pm$0.0013 & \textbf{0.2989$\pm$0.0009} & 0.0795$\pm$0.0004 & \textbf{0.0790$\pm$0.0003} & 0.5893$\pm$0.1996 & \textbf{0.5546$\pm$0.1765} \\
HyperIMTS & 0.5568$\pm$0.0179 & \textbf{0.5457$\pm$0.0156} & 0.2987$\pm$0.0006 & \textbf{0.2980$\pm$0.0005} & 0.0818$\pm$0.0008 & \textbf{0.0813$\pm$0.0007} & 0.4077$\pm$0.0191 & \textbf{0.3973$\pm$0.0165} \\

\bottomrule
\end{tabular}
}
\end{table*}

To evaluate the effectiveness of \undercal, we report forecasting results on four IMTS datasets with various models in Table \ref{tab:main_results_mse}, measured by the MSE metric. The lookback time window is 36 hours for MIMIC and PhysioNet, 3000 milliseconds for Human Activity, and 3 years for USHCN. We use the next 3 timestamps as forecast targets for MIMIC, PhysioNet, and USHCN, and use 300 milliseconds for Human Activity as forecast length, following the setting in existing works \cite{bilovs2021neural, de2019gru}. As can be seen, \undercal improves performance over most of the non-adaptive baselines, with particularly large gains on datasets exhibiting strong distribution shift. 

Specifically, on Human Activity dataset, we observe clear improvements across all models. For instance, the MSE of tPatchGNN decreases by 8.41\%, and that of Warpformer drops even more substantially by 21.84\%. This indicates that in scenarios with significant distribution shifts, our method delivers strong practical benefits and notable gains. On USHCN dataset, it improves the average MSE of Crossformer and GraFITi by 7.82\% and 5.89\%, respectively. On MIMIC dataset, \undercal reduces the MSE of mTAN by about 30.07\%, and also brings a 2.01\% improvement to HyperIMTS. The similar performance improvement is observed on the PhysioNet dataset as well. 
These results demonstrate that \undercal is a model-agnostic and effective solution for online irregular multivariate time series forecasting. 

\subsection{Ablation Study}
\label{sec:ablation-study}

 In order to verify the effectiveness of our designs, we evaluate the performance of \undercal and six variants on all benchmark datasets to analyze the contribution of each core component in our online adaptation framework. Specifically, 
(1) \textit{w/o GDC (Single Expert, Joint)} collapses the dual experts in the GDC into a single calibration expert, where all samples are updated jointly; 
(2) \textit{w/o GDC (Single Expert, Reliable)} updates only the reliable expert using low-uncertainty samples, while high-uncertainty samples bypass calibration; 
(3) \textit{w/o GDC (Single Expert, Unreliable)} updates only the unreliable expert using high-uncertainty samples, while reliable samples are not used for calibration; 
(4) \textit{w/o ARM (Random Triggering)} removes uncertainty-driven selective adaptation triggering and allows each batch to trigger adaptation with a fixed probability.
(5) \textit{w/o ARM (Random Allocating)} removes uncertainty-guided sample routing and randomly assigns samples to the two experts with equal probability;
(6) \textit{w/o All (Single Expert, Joint)} removes UE and uses a single calibration expert to adjust all samples per batch. The ablation results are summarized in Table~\ref{tab:ablation}. 
All ablation experiments are conducted under the same online setting and use identical source forecasters and hyperparameters unless specified.

As shown in Table~\ref{tab:ablation}, all modules and mechanisms in \undercal are necessary for effective online IMTS forecasting. The inferior performance of \textit{w/o GDC (Single Expert, Joint)} indicates that unified online updates are insufficient for handling samples with heterogeneous reliability. Results from \textit{w/o GDC (Single Expert, Reliable)} and \textit{w/o GDC (Single Expert, Unreliable)} further show that relying exclusively on either low or high uncertainty samples leads to suboptimal adaptation, highlighting the necessity of isolating update pathways. 
\textit{w/o ARM (Random Triggering)} exhibits unstable behavior or excessive update frequency, demonstrating that the UE and uncertainty-driven ARM are critical for balancing adaptation effectiveness and performance in online IMTS forecasting. 
The performance drop under \textit{w/o ARM (Random Allocating)} as well confirms that uncertainty-guided routing drives performance gains. 
Moreover, experiments on \textit{w/o All (Single Expert, Joint)} confirm that the UE, GDC and ARM together constitute the core mechanism for the effective operation of the system. The ARM ensures separate calibration and necessary updates, while the GDC performs targeted calibration for samples with different uncertainty levels. The effectiveness of the system relies on the synergistic operation of all components, none of which can be omitted.

\begin{table}
\centering
\caption{Ablation study of \undercal taking tPatchGNN as the source forecaster on four datasets under five seeds evaluated using MSE.}
\label{tab:ablation}
\resizebox{\linewidth}{!}{
\begin{tabular}{lcccc}
\toprule
Method Variant & MIMIC & PhysioNet & Human Activity & USHCN \\
\midrule
w/o GDC (Single Expert, Joint) & 0.4902 & 0.3070 & 0.1294 & 0.4549 \\
w/o GDC (Single Expert, Reliable) & 0.4937 & 0.3074 & 0.1333 & 0.4547 \\
w/o GDC (Single Expert, Unreliable) & 0.5001 & 0.3079 & 0.1360 & 0.4608 \\

\midrule
w/o ARM (Random Triggering) & 0.4915 & 0.3077 & 0.1371 & 0.4577 \\
w/o ARM (Random Allocating) & 0.4909 & 0.3069 & 0.1269 & 0.4567 
\\

\midrule
w/o All (Single Expert, Joint) &  0.4911 & 0.3070 & 0.1291 & 0.4539 \\

\midrule

\undercal & \textbf{0.4872} & \textbf{0.3066} & \textbf{0.1262} & \textbf{0.4538} \\

\bottomrule
\end{tabular}
}
\end{table}

\subsection{Parameter Study}
\label{sec:parameter-study}

\begin{figure}
    \centering
    \begin{subfigure}[b]{0.49\linewidth}
        \centering
        \includegraphics[width=\linewidth]{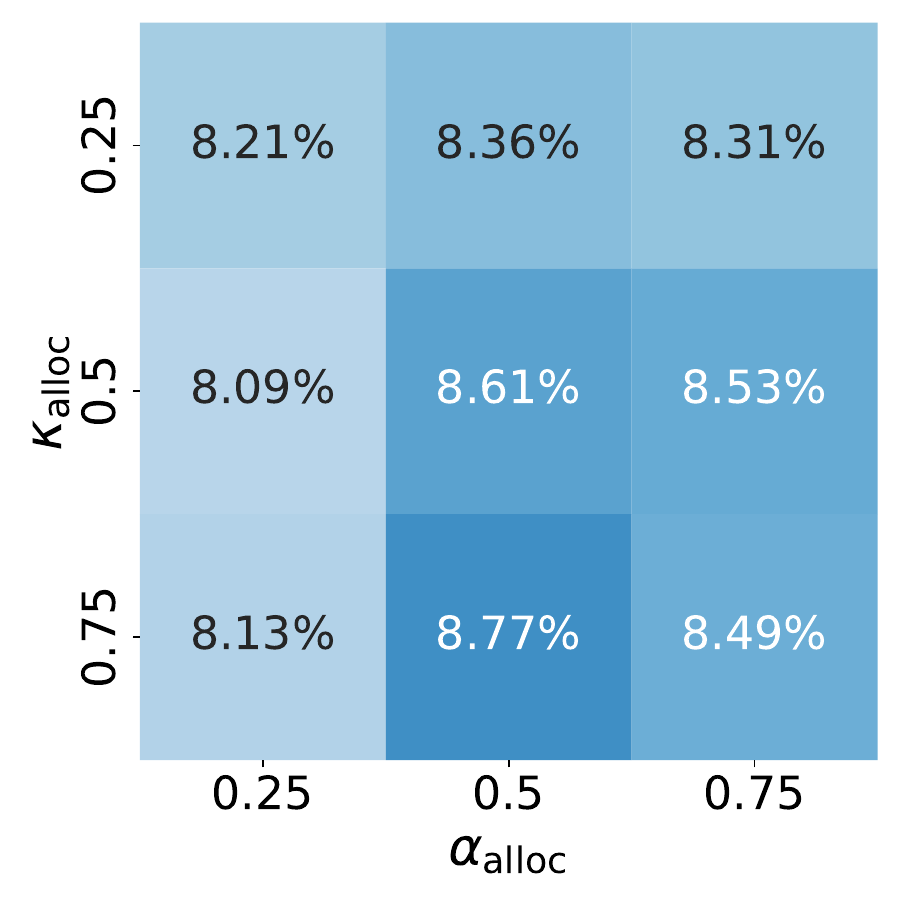}
        \caption{Allocation threshold}
        \label{fig:parameter_study_allocation}
    \end{subfigure}
    \hfill
    \begin{subfigure}[b]{0.49\linewidth}
        \centering
        \includegraphics[width=\linewidth]{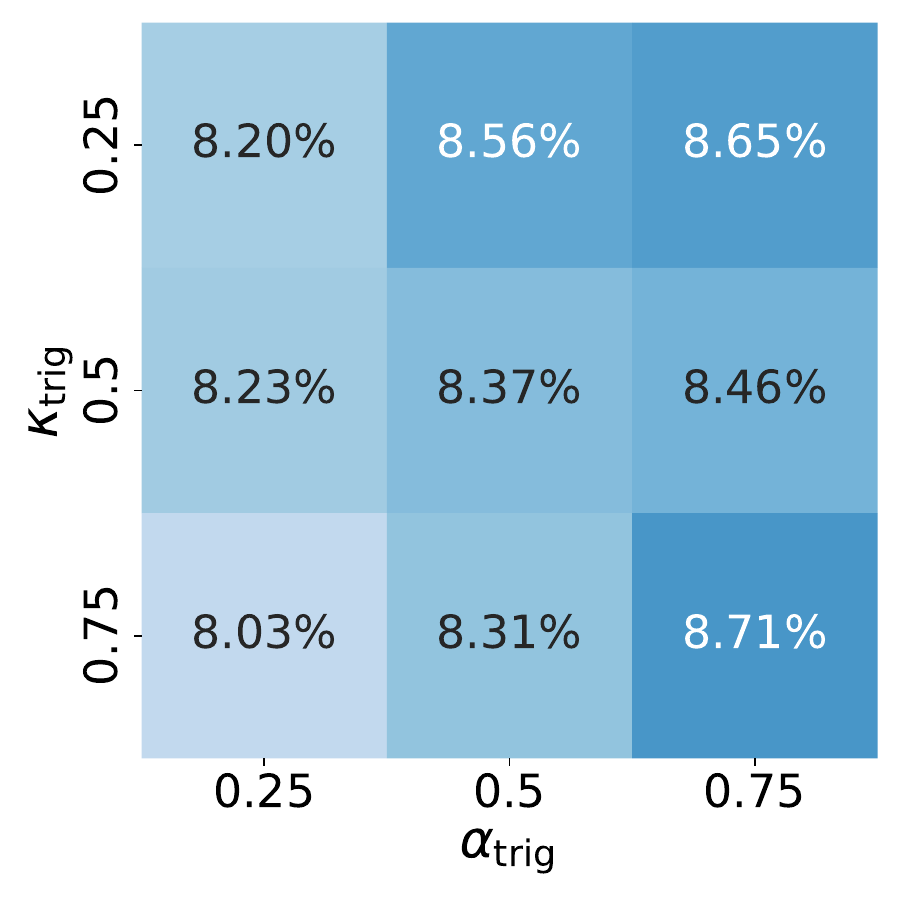}
        \caption{Trigger threshold}
        \label{fig:parameter_study_trigger}
    \end{subfigure}

    \caption{Parameter Study of Human Activity on tPatchGNN.}
    \label{fig:parameter_study}
\end{figure}

\undercal involves four key hyperparameters that require experimental analysis, including the smoothing factor $\alpha_\mathrm{alloc}$ for the allocation threshold $\tau_\mathrm{alloc}$, the smoothing factor $\alpha_\mathrm{trig}$ for the trigger threshold $\tau_\mathrm{trig}$, and the positive coefficients $\kappa_\mathrm{alloc}$ and $\kappa_\mathrm{trig}$, which control the sensitivity of allocation and triggering, respectively. The parameter experiment results of Human Activity on tPatchGNN are summarized in Figure \ref{fig:parameter_study}. We provide the average performance improvement of different combinations of $\alpha$ and $\kappa$ for allocation and trigger thresholds, indicating that the performance is insensitive across parameter choices. However, we also observe that different parameter combinations lead to varied update rates. The selected parameter set mentioned in Section \ref{sec:experimental-setup} maintains strong predictive performance while significantly reducing the model update frequency by an average of approximately 15\%. It helps us to achieve an effective balance between performance and efficiency.

\subsection{Lookback Length and Horizon Study}
\label{sec:varying-lookback}

To further assess the robustness of our method across varying lookback lengths and forecasting horizons, we provide the results in Table \ref{tab:vary_window_results_mse}. For varying lookback lengths, we keep the same forecast horizons in Section \ref{sec:main-results} and set the lookback window lengths to: (1) MIMIC: 12, 24, and 36 hours; (2) PhysioNet: 12, 24, and 36 hours; (3) Human Activity: 1200, 2100, and 3000 milliseconds; (4) USHCN: 1, 2, and 3 years. As can be seen, \undercal generally improves model performance across different lookback lengths, demonstrating its ability to leverage historical information and mitigate distribution shifts. 
For varying forecast horizons, we follow the same settings in Section \ref{sec:main-results} as well. For MIMIC and PhysioNet, the forecast horizon is set to 12 hours. For Human Activity and USHCN, the forecast horizon is 1000 milliseconds and 1 year, respectively. We find that under the more challenging setting of extended forecast horizons, \undercal shows slower performance degradation and greater stability compared to baselines, which proves its particular strength in handling distribution shifts in long-term forecasting.

In summary, \undercal is not only robust across various historical contexts but also exhibits critical advantages in long-horizon scenarios, providing models with continuous, adaptive calibration that significantly enhances the accuracy and reliability of online irregular multivariate time series forecasting.

\begin{table}
\centering
\caption{Test MSE on irregular multivariate time series forecasting datasets varying lookback window lengths and forecast horizons under five seeds. Lower MSE indicates better performance. Best results are highlighted in bold.}
\label{tab:vary_window_results_mse}
\resizebox{\linewidth}{!}{
\begin{tabular}{cccccccccc}
\toprule
Models & \multirow{2}{*}{Length} & \multicolumn{2}{c}{tPatchGNN} & \multicolumn{2}{c}{HyperIMTS} & \multicolumn{2}{c}{GraFITi} & \multicolumn{2}{c}{Warpformer}\\
\cmidrule(lr){3-4} \cmidrule(lr){5-6} \cmidrule(lr){7-8} \cmidrule(lr){9-10}
+\undercal &  & \rxmark & \gcmark & \rxmark & \gcmark & \rxmark & \gcmark & \rxmark & \gcmark\\
\midrule
\multirow{4}{*}{MIMIC}
 & 24 $\rightarrow$ 3 & 0.8311 & \textbf{0.8113} & 0.8115 & \textbf{0.8100} & 0.7289 & \textbf{0.7207} & 0.7561 & \textbf{0.7400} \\
 & 48 $\rightarrow$ 3 & 0.4653 & \textbf{0.4574} & 0.5206 & \textbf{0.5166} & 0.4694 & \textbf{0.4688} & 0.4900 & \textbf{0.4879} \\
 & 72 $\rightarrow$ 3 & 0.4943 & \textbf{0.4872} & 0.5568 & \textbf{0.5457} & 0.4531 & \textbf{0.4482} & 0.4722 & \textbf{0.4683} \\
 & 72 $\rightarrow$ 24 & 0.5814 & \textbf{0.5801} & 0.5805 & \textbf{0.5782} & 0.5383 & \textbf{0.5339} & 0.5532 & \textbf{0.5551} \\

\midrule

\multirow{4}{*}{PhysioNet}
  & 12 $\rightarrow$ 3 & 0.3447 & \textbf{0.3417} & 0.3272 & \textbf{0.3263} & 0.3310 & \textbf{0.3302} & 0.3372 & \textbf{0.3355} \\
  & 24 $\rightarrow$ 3 & 0.3257 & \textbf{0.3230} & 0.3137 & \textbf{0.3122} & 0.3141 & \textbf{0.3133} & 0.3202 & \textbf{0.3184} \\
  & 36 $\rightarrow$ 3 & 0.3080 & \textbf{0.3067} & 0.2987 & \textbf{0.2980} & 0.2996 & \textbf{0.2989} & 0.3040 & \textbf{0.3026} \\
  & 36 $\rightarrow$ 12 & 0.3796 & \textbf{0.3779} & 0.3698 & \textbf{0.3691} & 0.3734 & \textbf{0.3725} & 0.3740 & \textbf{0.3733} \\

\midrule
  
\multirow{4}{*}{Human Activity}
  & 1200 $\rightarrow$ 300 & 0.1215 & \textbf{0.1106} & 0.0701 & \textbf{0.0695} & 0.0680 & \textbf{0.0672} & 0.4474 & \textbf{0.3332} \\
  & 2100 $\rightarrow$ 300 & 0.0957 & \textbf{0.0902} & 0.0716 & \textbf{0.0711} & 0.0696 & \textbf{0.0691} & 0.4171 & \textbf{0.3593} \\
  & 3000 $\rightarrow$ 300 & 0.1404 & \textbf{0.1287} & 0.0818 & \textbf{0.0814} & 0.0795 & \textbf{0.0790} & 0.3942 & \textbf{0.3105} \\
  & 3000 $\rightarrow$ 1000 & 0.1550 & \textbf{0.1408} & 0.1005 & \textbf{0.1004} & 0.0993 & \textbf{0.0992} & 0.4643 & \textbf{0.3455} \\

\midrule
  
\multirow{4}{*}{USHCN}
  & 50 $\rightarrow$ 3 & 0.5166 & \textbf{0.5146} & 0.5366 & \textbf{0.5311} & 1.5995 & \textbf{1.5878} & 1.6039 & \textbf{1.5993} \\
  & 100 $\rightarrow$ 3 & 0.9845 & \textbf{0.9797} & 0.9667 & \textbf{0.9644} & 0.2920 & \textbf{0.2836} & 0.2937 & \textbf{0.2859} \\
  & 150 $\rightarrow$ 3 & 0.4589 & \textbf{0.4538} & 0.4077 & \textbf{0.3977} & 0.5893 & \textbf{0.5638} & 0.4338 & \textbf{0.4294} \\
  & 150 $\rightarrow$ 50 & 0.8473 & \textbf{0.7592} & 0.7940 & \textbf{0.7643} & 0.7719 & \textbf{0.7361} & 0.6264 & \textbf{0.6240} \\

\bottomrule
\end{tabular}
}
\end{table}

\subsection{Comparison with Online and TTA Methods}
\label{sec:comparison-with-Online}

To further evaluate the effectiveness of \undercal, we compare against several representative online learning and TTA methods, including OneNet, FSNet, D3A, and TAFAS. Among them, OneNet and FSNet tightly couple online updating mechanisms with proprietary forecasting architectures, and we further investigate whether \undercal can enhance their online adaptation capability. Specifically, for OneNet, we compare the variants with and without online learning, as well as the integration with \undercal. For FSNet, we compare the vanilla FSNet, FSNet+D3A, and FSNet+\undercal. In addition, we conduct plug-in adaptation experiments on HyperIMTS using D3A and TAFAS as representative methods. D3A triggers model updates when distribution drift is detected and performs adaptation using accumulated historical data together with additional noise perturbation, while TAFAS performs test-time adaptation through adaptive periodicity-aware mechanisms. Results in Tables~\ref{tab:comparison-with-Online1}, Table~\ref{tab:comparison-with-Online2}, and Table~\ref{tab:comparison-with-Online3} show that \undercal consistently achieves the best performance across different datasets and forecasting backbones.

Notably, conventional online learning and TTA methods such as D3A and TAFAS may even lead to performance degradation in IMTS settings, particularly on HyperIMTS. This suggests that directly transferring adaptation strategies developed for regularly sampled time series to irregular observation scenarios may introduce unstable updates due to asynchronous observations and dynamically evolving missingness patterns.

In contrast, \undercal effectively improves online forecasting robustness through uncertainty-aware calibration and dual-expert adaptation, consistently reducing forecasting errors.

\begin{table}
\caption{Test MSE and MAE of OneNet under different online learning settings on the Human Activity and USHCN datasets. Best results are highlighted in bold.}
\footnotesize
\begin{tabular*}{\linewidth}{@{\extracolsep{\fill}}ccccc}
\toprule
\multirow{2}{*}{Dataset} & \multirow{2}{*}{Metric}
& \multicolumn{3}{c}{OneNet}\\
\cmidrule(lr){3-5}
&  & \rxmark & +Online & +\undercal\\
\midrule
\multirow{2}{*}{Human Activity}
& MSE & 0.3360 & 0.3151 & \textbf{0.2780} \\
& MAE & 0.4645 & 0.4409 & \textbf{0.3830} \\
\midrule
\multirow{2}{*}{USHCN}
& MSE & 0.7390 & 0.7130 & \textbf{0.5910} \\
& MAE & 0.4533 & 0.4540 & \textbf{0.3835} \\
\bottomrule
\end{tabular*}
\label{tab:comparison-with-Online1}
\end{table}

\begin{table}
\caption{Test MSE and MAE of FSNet under different online adaptation strategies on the Human Activity and USHCN datasets. Best results are highlighted in bold.}
\footnotesize
\begin{tabular*}{\linewidth}{@{\extracolsep{\fill}}ccccc}
\toprule
\multirow{2}{*}{Dataset} & \multirow{2}{*}{Metric}
& \multicolumn{3}{c}{FSNet}\\
\cmidrule(lr){3-5}
&  & \rxmark & +D3A & +\undercal\\
\midrule
\multirow{2}{*}{Human Activity}
& MSE & 0.3960 & 0.3911 & \textbf{0.3694} \\
& MAE & 0.4595 & 0.4566 & \textbf{0.4441} \\
\midrule
\multirow{2}{*}{USHCN}
& MSE & 0.8059 & 0.8017 & \textbf{0.7449} \\
& MAE & 0.4198 & 0.4162 & \textbf{0.3929} \\
\bottomrule
\end{tabular*}
\label{tab:comparison-with-Online2}
\end{table}

\begin{table}
\caption{Test MSE and MAE of plug-in online and TTA methods on HyperIMTS over the Human Activity and USHCN datasets. Best results are highlighted in bold.}
\footnotesize
\begin{tabular*}{\linewidth}{@{\extracolsep{\fill}}cccccc}
\toprule
\multirow{2}{*}{Dataset} & \multirow{2}{*}{Metric}
& \multicolumn{4}{c}{HyperIMTS} \\
\cmidrule(lr){3-6}
&  & \rxmark & +D3A & +TAFAS & +\undercal \\
\midrule
\multirow{2}{*}{Human Activity}
& MSE & 0.0818 & 0.2362 & 0.2505 & \textbf{0.0813} \\
& MAE & 0.1877 & 0.3576 & 0.3332 & \textbf{0.1869} \\
\midrule
\multirow{2}{*}{USHCN}
& MSE & 0.3809 & 0.4418 & 0.6376 & \textbf{0.3754} \\
& MAE & 0.3448 & 0.3821 & 0.5420 & \textbf{0.3389} \\
\bottomrule
\end{tabular*}
\label{tab:comparison-with-Online3}
\end{table}

\subsection{Case Study}
\label{sec:case-study}

\begin{figure}[t]
    \centering
    \begin{subfigure}[b]{0.49\linewidth}
        \centering
        \includegraphics[width=\linewidth]{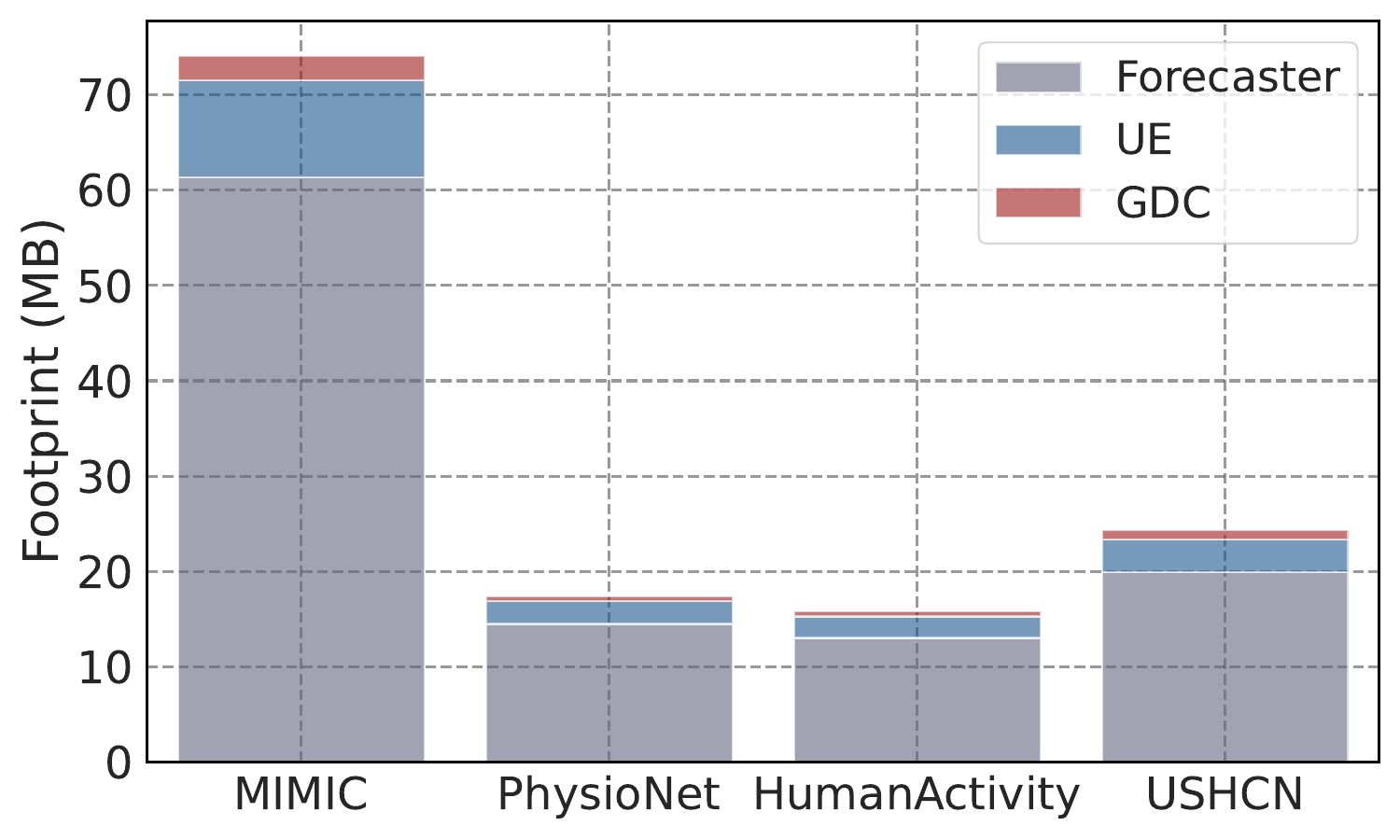}
        \caption{GPU footprint across four datasets.}
        \label{fig:footprint}
    \end{subfigure}
    \hfill
    \begin{subfigure}[b]{0.49\linewidth}
        \centering
        \includegraphics[width=\linewidth]{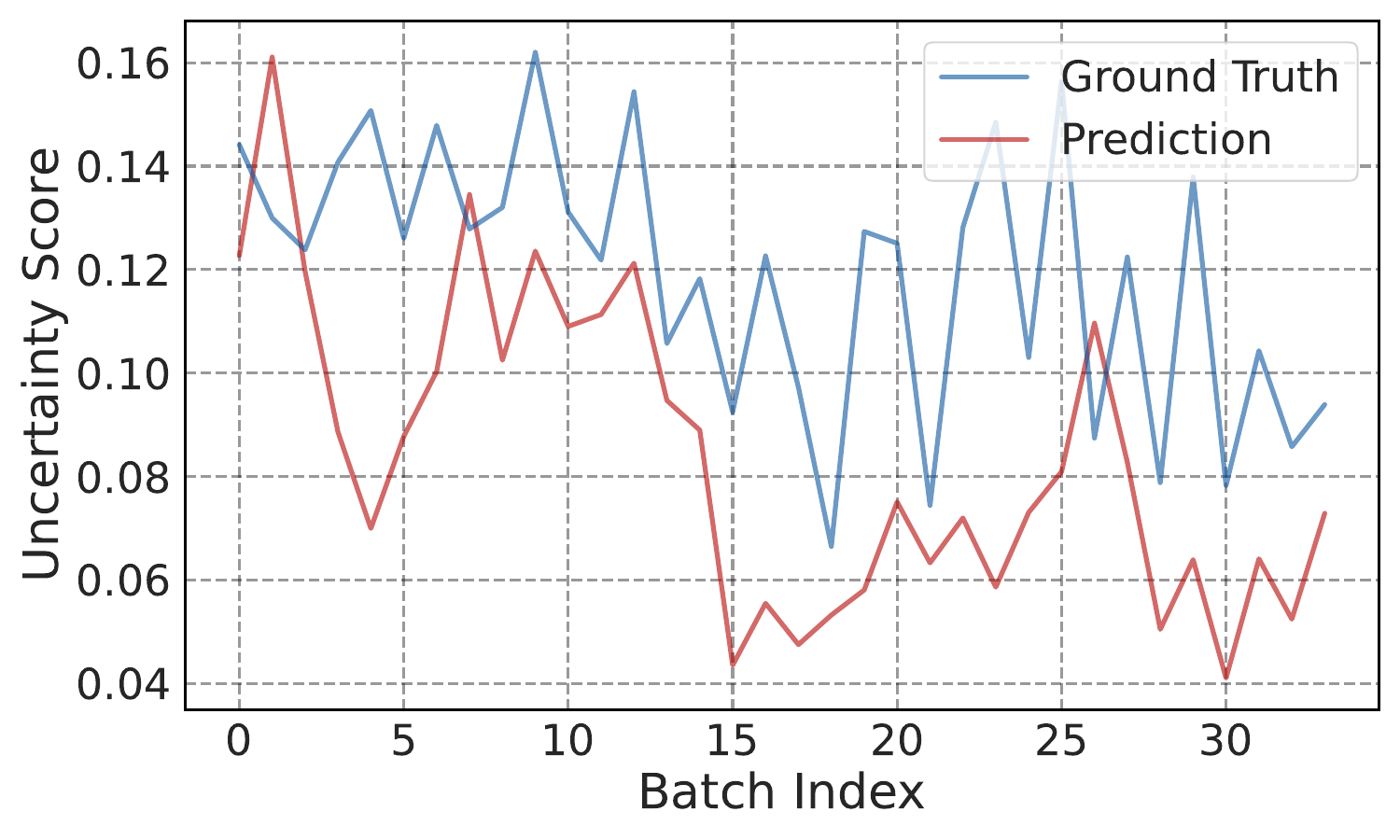}
        \caption{Uncertainty score estimations on Human Activity.}
        \label{fig:uncertainty_score_humanactivity}
    \end{subfigure}

    \hfill
    \begin{subfigure}[b]{0.49\linewidth}
        \centering
        \includegraphics[width=\linewidth]{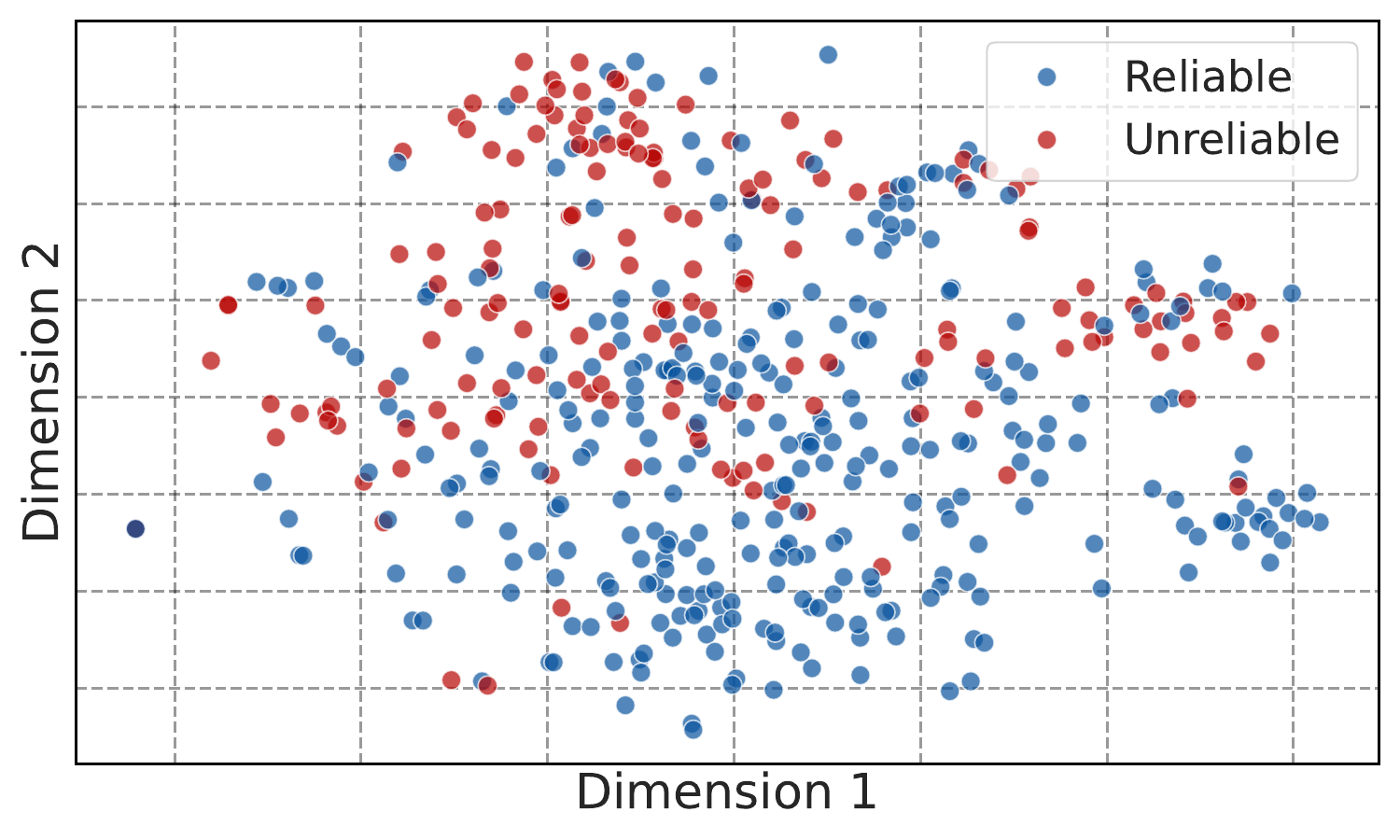}
        \caption{Allocated sample distribution on PhysioNet.}
        \label{fig:sample_distribution_P12}
    \end{subfigure}
    \hfill
    \begin{subfigure}[b]{0.49\linewidth}
        \centering
        \includegraphics[width=\linewidth]{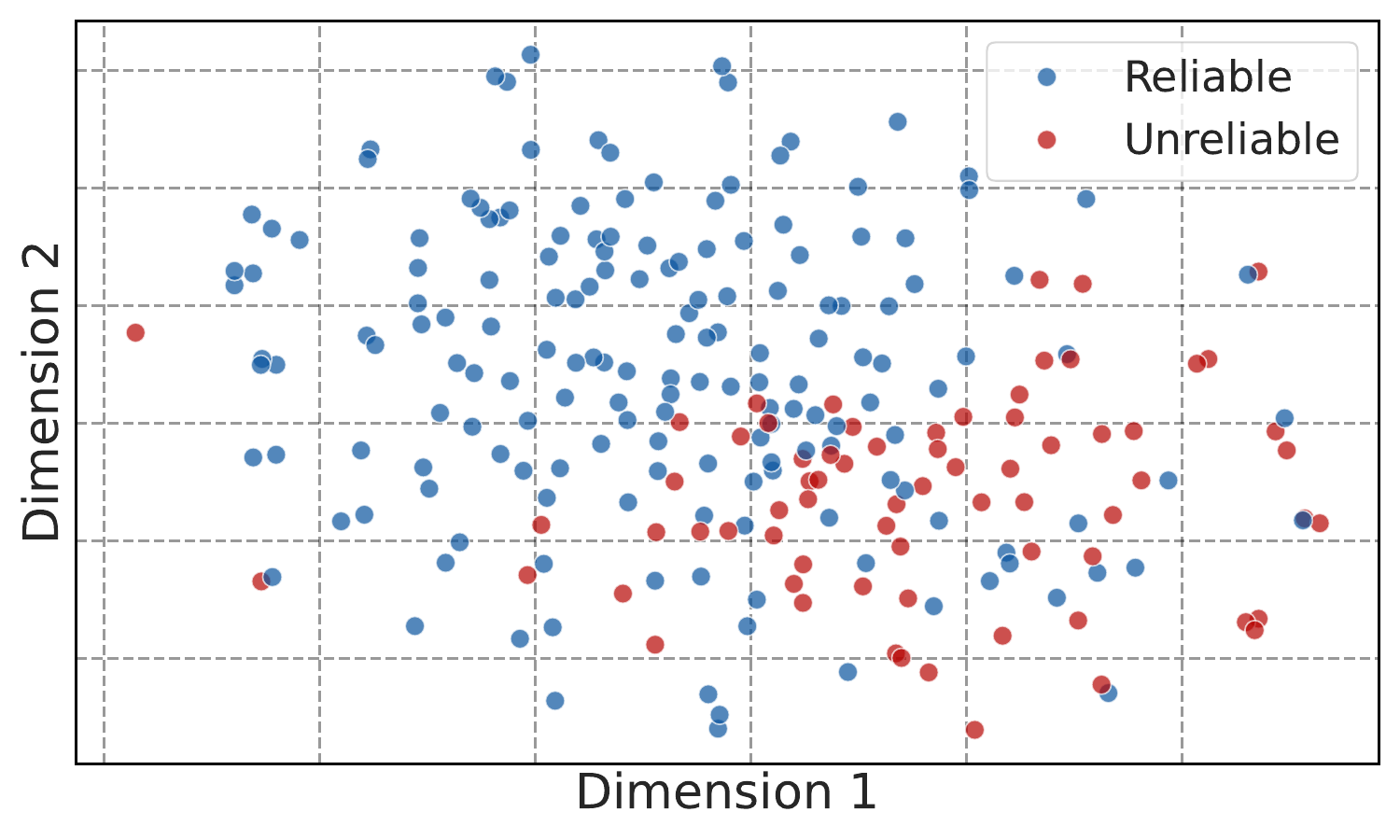}
        \caption{Allocated sample distribution on USHCN.}
        \label{fig:sample_distribution_USHCN}
    \end{subfigure}

    \caption{Case study on tPatchGNN.}
    \label{fig:case_study}
\end{figure}

\paragraph{GPU Footprint of \undercal.}
We analyze the GPU footprint of \undercal with a batch size of one, and provide the results on tPatchGNN in Figure \ref{fig:footprint}. As shown, integrating \undercal leads to only marginal increases in GPU memory usage. This demonstrates that our method achieves resource-efficient online learning, making it practical for deployment alongside large-scale forecasters.

\paragraph{Inference Latency of \undercal.}

As shown in Table \ref{tab:inference-time}, although \undercal introduces additional adaptation operations, 
the batch-level inference latency remains within 0.1–0.4s, which is acceptable in real scenarios.
This indicates that the limited computational overhead is well offset by performance gains, demonstrating a favorable trade-off between adaptive capability and inference efficiency.

\paragraph{Consistency of Uncertainty Scores.}
To validate the effectiveness and accuracy of UE, we compare the estimated uncertainty scores $\hat{u}$ against the ground truth, \textit{i.e.}, prediction errors of the calibrated outputs $u$. Taking the results on tPatchGNN over Human Activity dataset as an example, shown in Figure \ref{fig:uncertainty_score_humanactivity}, we observe that the trend of UE outputs aligns well with the ground truth. It shows that the UE provides consistent estimations of uncertainty scores. Meanwhile, we also observe that the estimations are slightly lower compared to the ground truth. We believe that the estimation bias can be effectively alleviated by the adaptive design of thresholds in ARM, which further ensures the calibration stability and reliability.

\paragraph{Sample Distributions in Dual-Expert Architecture.}
To further verify the effectiveness of dual-expert calibration architecture in GDC, we use t-SNE \cite{maaten2008visualizing} to visualize the expert-wise sample distributions on tPatchGNN over the PhysioNet and USHCN datasets, as shown in Figure \ref{fig:sample_distribution_P12} and \ref{fig:sample_distribution_USHCN}. We observe clear distributional shifts between the samples processed by the two experts in both figures, indicating that our dual-expert calibration architecture and adaptive routing module can effectively guide and learn samples with different uncertainty scores, thereby assisting in stable calibration.

\begin{table}[t]
\centering
\caption{Batch-level inference time (ms) for tPatchGNN and HyperIMTS with and without \undercal on four datasets.}
\footnotesize
\label{tab:inference-time}
    \begin{tabular*}{\linewidth}{@{\extracolsep{\fill}}ccccc}
    \toprule
    Dataset 
    & \multicolumn{2}{c}{tPatchGNN}
    & \multicolumn{2}{c}{HyperIMTS} \\ 
    \cmidrule(lr){2-3} \cmidrule(lr){4-5}
    +\undercal & \rxmark & \gcmark & \rxmark & \gcmark \\
    \midrule
    MIMIC         & 58.50  & 176.82 & 49.84 & 139.83 \\
    PhysioNet     & 11.87  & 112.94 & 43.38 & 243.26 \\
    Human Activity & 163.98 & 126.56 & 80.29 & 385.73 \\
    USHCN         & 48.39  & 83.83 & 56.09 & 188.45 \\
    \bottomrule
    \end{tabular*}
\end{table}

\section{Conclusion}
\label{sec:conclusion}

This study addresses the distribution shifts problem in online irregular multivariate time series forecasting by proposing \undercal, a model-agnostic framework driven by the synergistic collaboration of an uncertainty estimator, a dual-expert calibration module, and an adaptive routing module. The framework dynamically guides sample allocation and update decisions based on uncertainty, and employs isolated dual-expert calibration for samples of differing reliability, enabling efficient and stable online adaptation without updating the source forecaster. Extensive experiments show that \undercal delivers consistent performance gains across various source forecasters, with pronounced advantages under significant distribution shifts and in long-horizon forecasting. Ablation studies confirm the necessity of each core component.

\begin{acks}
This work was supported by the Beijing Natural Science Foundation (No. 4242046, No. L259051) and the Natural Science Foundation of Hebei Province (No. F2025105018).
\end{acks}

\clearpage
\bibliographystyle{ACM-Reference-Format}
\bibliography{sample-base}

\appendix

\begin{table*}
\centering
\caption{Test MAE (mean $\pm$ std) on irregular multivariate time series forecasting datasets with and without \undercal across various architectures under five seeds. Lower MAE indicates better performance. Best results are highlighted in bold.}
\label{tab:main_results_mae}
\resizebox{\linewidth}{!}{
\begin{tabular}{ccccccccc}
\toprule
Models & \multicolumn{2}{c}{MIMIC} & \multicolumn{2}{c}{PhysioNet} & \multicolumn{2}{c}{Human Activity} & \multicolumn{2}{c}{USHCN}\\
\cmidrule(lr){2-3} \cmidrule(lr){4-5} \cmidrule(lr){6-7} \cmidrule(lr){8-9}
+\undercal & \rxmark & \gcmark & \rxmark & \gcmark & \rxmark & \gcmark & \rxmark & \gcmark\\
\midrule
FEDformer & 0.5646$\pm$0.0021 & \textbf{0.5162$\pm$0.0038} & 0.4548$\pm$0.0010 & \textbf{0.4316$\pm$0.0008} & 0.4833$\pm$0.0110 & \textbf{0.4694$\pm$0.0120} & 0.3854$\pm$0.0069 & \textbf{0.3723$\pm$0.0099} \\
FreTS & 0.5113$\pm$0.0012 & \textbf{0.4937$\pm$0.0010} & 0.4363$\pm$0.0015 & \textbf{0.4223$\pm$0.0015} & 0.2418$\pm$0.0011 & \textbf{0.2382$\pm$0.0010} & 0.4433$\pm$0.0136 & \textbf{0.4180$\pm$0.0112} \\
BigST & 0.4756$\pm$0.0021 & \textbf{0.4749$\pm$0.0016} & 0.4216$\pm$0.0034 & \textbf{0.4114$\pm$0.0021} & 0.4816$\pm$0.0172 & \textbf{0.3986$\pm$0.0124} & 0.3817$\pm$0.0161 & \textbf{0.3774$\pm$0.0144} \\
iTransformer & 0.4916$\pm$0.0016 & \textbf{0.4845$\pm$0.0045} & 0.4384$\pm$0.0023 & \textbf{0.4232$\pm$0.0012} & 0.2395$\pm$0.0027 & \textbf{0.2299$\pm$0.0012} & 0.3729$\pm$0.0093 & \textbf{0.3636$\pm$0.0026} \\
Ada-MSHyper & 0.5233$\pm$0.0044 & \textbf{0.5123$\pm$0.0043} & 0.4363$\pm$0.0005 & \textbf{0.4244$\pm$0.0049} & 0.3916$\pm$0.0050 & \textbf{0.3562$\pm$0.0170} & 0.3911$\pm$0.0099 & \textbf{0.3807$\pm$0.0066} \\
PatchTST & 0.5139$\pm$0.0017 & \textbf{0.4898$\pm$0.0068} & \textbf{0.4306$\pm$0.0010} & 0.4409$\pm$0.0148 & 0.2457$\pm$0.0037 & \textbf{0.2426$\pm$0.0039} & 0.4107$\pm$0.0104 & \textbf{0.3958$\pm$0.0094} \\
Reformer & \textbf{0.6013$\pm$0.0016} & 0.6284$\pm$0.0433 & \textbf{0.4573$\pm$0.0008} & 0.4576$\pm$0.0048 & 0.8772$\pm$0.0312 & \textbf{0.6886$\pm$0.1016} & 0.3961$\pm$0.0024 & \textbf{0.3891$\pm$0.0029} \\
Informer & \textbf{0.5692$\pm$0.0044} & 0.6027$\pm$0.0242 & 0.5147$\pm$0.0019 & \textbf{0.5084$\pm$0.0056} & 0.5308$\pm$0.0184 & \textbf{0.4854$\pm$0.0142} & \textbf{0.3730$\pm$0.0039} & 0.3840$\pm$0.0080 \\
Crossformer & \textbf{0.4534$\pm$0.0014} & 0.4553$\pm$0.0012 & 0.3996$\pm$0.0021 & \textbf{0.3988$\pm$0.0014} & 0.3695$\pm$0.0189 & \textbf{0.3489$\pm$0.0184} & 0.3883$\pm$0.0139 & \textbf{0.3594$\pm$0.0095} \\
\midrule
PrimeNet & 0.6550$\pm$0.0001 & \textbf{0.6414$\pm$0.0010} & 0.6891$\pm$0.0001 & \textbf{0.6784$\pm$0.0017} & 1.6478$\pm$0.0011 & \textbf{1.5154$\pm$0.0140} & 0.6147$\pm$0.0021 & \textbf{0.5272$\pm$0.0046} \\
SeFT & 0.6482$\pm$0.0011 & \textbf{0.6316$\pm$0.0041} & 0.6883$\pm$0.0015 & \textbf{0.6599$\pm$0.0054} & 1.0711$\pm$0.0193 & \textbf{0.8976$\pm$0.1195} & 0.4893$\pm$0.0021 & \textbf{0.4424$\pm$0.0071} \\
mTAN & 0.7852$\pm$0.0574 & \textbf{0.6365$\pm$0.0194} & 0.4668$\pm$0.0049 & \textbf{0.4600$\pm$0.0035} & 0.4396$\pm$0.0092 & \textbf{0.4299$\pm$0.0103} & 0.4472$\pm$0.0242 & \textbf{0.4035$\pm$0.0118} \\
NeuralFlows & 0.6236$\pm$0.0096 & \textbf{0.6039$\pm$0.0109} & 0.4789$\pm$0.0044 & \textbf{0.4753$\pm$0.0047} & 0.5117$\pm$0.0222 & \textbf{0.4917$\pm$0.0108} & 0.4088$\pm$0.0060 & \textbf{0.3969$\pm$0.0083} \\
GNeuralFlow & 0.6448$\pm$0.0068 & \textbf{0.6314$\pm$0.0030} & 0.6504$\pm$0.0222 & \textbf{0.6119$\pm$0.0102} & 0.4822$\pm$0.0071 & \textbf{0.4725$\pm$0.0080} & 0.4089$\pm$0.0073 & \textbf{0.4024$\pm$0.0103} \\
CRU & 0.6050$\pm$0.0044 & \textbf{0.5940$\pm$0.0028} & 0.6052$\pm$0.0026 & \textbf{0.5855$\pm$0.0023} & 0.4661$\pm$0.0323 & \textbf{0.4318$\pm$0.0193} & 0.4083$\pm$0.0062 & \textbf{0.3993$\pm$0.0064} \\
GRU-D & 0.5322$\pm$0.0052 & \textbf{0.5275$\pm$0.0047} & 0.4281$\pm$0.0011 & \textbf{0.4268$\pm$0.0005} & 0.4488$\pm$0.0008 & \textbf{0.4456$\pm$0.0014} & 0.3892$\pm$0.0134 & \textbf{0.3787$\pm$0.0098} \\
Warpformer & \textbf{0.4068$\pm$0.0066} & 0.4077$\pm$0.0024 & 0.3656$\pm$0.0018 & \textbf{0.3648$\pm$0.0008} & 0.3978$\pm$0.0198 & \textbf{0.3542$\pm$0.0169} & 0.3564$\pm$0.0237 & \textbf{0.3405$\pm$0.0147} \\
tPatchGNN & 0.4281$\pm$0.0122 & \textbf{0.4269$\pm$0.0094} & 0.3682$\pm$0.0011 & \textbf{0.3672$\pm$0.0009} & 0.2502$\pm$0.0885 & \textbf{0.2359$\pm$0.0726} & 0.3547$\pm$0.0252 & \textbf{0.3479$\pm$0.0230} \\
Hi-Patch & 0.4470$\pm$0.0090 & \textbf{0.4426$\pm$0.0053} & 0.3999$\pm$0.0030 & \textbf{0.3962$\pm$0.0017} & 0.8279$\pm$0.2170 & \textbf{0.8109$\pm$0.2181} & 0.3396$\pm$0.0076 & \textbf{0.3388$\pm$0.0076} \\
GraFITi & 0.3958$\pm$0.0015 & \textbf{0.3945$\pm$0.0026} & 0.3611$\pm$0.0017 & \textbf{0.3604$\pm$0.0011} & 0.1773$\pm$0.0012 & \textbf{0.1764$\pm$0.0005} & 0.3444$\pm$0.0137 & \textbf{0.3366$\pm$0.0156} \\
HyperIMTS & 0.4697$\pm$0.0108 & \textbf{0.4640$\pm$0.0099} & 0.3620$\pm$0.0012 & \textbf{0.3612$\pm$0.0004} & 0.1877$\pm$0.0019 & \textbf{0.1869$\pm$0.0019} & 0.3453$\pm$0.0096 & \textbf{0.3330$\pm$0.0068} \\
\bottomrule
\end{tabular}
}
\end{table*}

\begin{table*}
\centering
\caption{Test MAE (mean $\pm$ std) on irregular multivariate time series forecasting datasets varying lookback window lengths and forecast horizons under five seeds. Lower MSE indicates better performance. Best results are highlighted in bold.}
\label{tab:vary_window_results_mae}
\resizebox{\linewidth}{!}{
\begin{tabular}{cccccccccc}
\toprule
Models & \multirow{2}{*}{Length} & \multicolumn{2}{c}{tPatchGNN} & \multicolumn{2}{c}{HyperIMTS} & \multicolumn{2}{c}{GraFITi} & \multicolumn{2}{c}{Warpformer}\\
\cmidrule(lr){3-4} \cmidrule(lr){5-6} \cmidrule(lr){7-8} \cmidrule(lr){9-10}
+\undercal &  & \rxmark & \gcmark & \rxmark & \gcmark & \rxmark & \gcmark & \rxmark & \gcmark\\
\midrule
\multirow{4}{*}{MIMIC}
& 24 $\rightarrow$ 3 & 0.5861$\pm$0.0021  & \textbf{0.5782$\pm$0.0019}  & 0.5511$\pm$0.0099  & \textbf{0.5492$\pm$0.0098}  & 0.5388$\pm$0.0033  & \textbf{0.5336$\pm$0.0028}  & 0.5522$\pm$0.0013  & \textbf{0.5454$\pm$0.0030}  \\
 & 48 $\rightarrow$ 3 & 0.4111$\pm$0.0054  & \textbf{0.4081$\pm$0.0030}  & 0.4491$\pm$0.0077  & \textbf{0.4453$\pm$0.0061}  & 0.4111$\pm$0.0010  & \textbf{0.4096$\pm$0.0004}  & 0.4212$\pm$0.0037  & \textbf{0.4202$\pm$0.0017}  \\
 & 72 $\rightarrow$ 3 & 0.4281$\pm$0.0122  & \textbf{0.4269$\pm$0.0094}  & 0.4697$\pm$0.0108  & \textbf{0.4640$\pm$0.0099}  & 0.3958$\pm$0.0015  & \textbf{0.3945$\pm$0.0026}  & \textbf{0.4068$\pm$0.0066}  & 0.4077$\pm$0.0024  \\
 & 72 $\rightarrow$ 24 & \textbf{0.4647$\pm$0.0050}  & 0.4667$\pm$0.0050  & 0.4738$\pm$0.0030  & \textbf{0.4724$\pm$0.0028}  & 0.4439$\pm$0.0065  & \textbf{0.4413$\pm$0.0057}  & \textbf{0.4476$\pm$0.0012}  & 0.4508$\pm$0.0014  \\

\midrule

\multirow{4}{*}{PhysioNet}
& 12 $\rightarrow$ 3 & 0.2434$\pm$0.0557  & \textbf{0.2293$\pm$0.0436}  & 0.1713$\pm$0.0030  & \textbf{0.1698$\pm$0.0025}  & 0.1615$\pm$0.0017  & \textbf{0.1595$\pm$0.0025}  & 0.4249$\pm$0.0322  & \textbf{0.3666$\pm$0.0213}  \\
& 24 $\rightarrow$ 3 & 0.2080$\pm$0.0497  & \textbf{0.1990$\pm$0.0404}  & 0.1732$\pm$0.0016  & \textbf{0.1718$\pm$0.0016}  & 0.1647$\pm$0.0016  & \textbf{0.1632$\pm$0.0017}  & 0.4104$\pm$0.0192  & \textbf{0.3810$\pm$0.0273}  \\
& 36 $\rightarrow$ 3 & 0.2566$\pm$0.0899  & \textbf{0.2417$\pm$0.0755}  & 0.1877$\pm$0.0019  & \textbf{0.1871$\pm$0.0020}  & 0.1773$\pm$0.0012  & \textbf{0.1764$\pm$0.0005}  & 0.3978$\pm$0.0198  & \textbf{0.3551$\pm$0.0184}  \\
& 36 $\rightarrow$ 12 & 0.2683$\pm$0.0912  & \textbf{0.2512$\pm$0.0713}  & 0.2082$\pm$0.0020  & \textbf{0.2083$\pm$0.0021}  & 0.1979$\pm$0.0010  & \textbf{0.1979$\pm$0.0009}  & 0.4390$\pm$0.0336  & \textbf{0.3810$\pm$0.0198}  \\

\midrule
  
\multirow{4}{*}{Human Activity}
& 1200 $\rightarrow$ 300 & 0.4057$\pm$0.0164  & \textbf{0.4018$\pm$0.0159}  & 0.4140$\pm$0.0301  & \textbf{0.4097$\pm$0.0291}  & 0.4576$\pm$0.0093  & \textbf{0.4548$\pm$0.0080}  & 0.4623$\pm$0.0081  & \textbf{0.4579$\pm$0.0093}  \\
& 2100 $\rightarrow$ 300 & 0.4651$\pm$0.0095  & \textbf{0.4569$\pm$0.0108}  & 0.4561$\pm$0.0117  & \textbf{0.4539$\pm$0.0116}  & 0.2872$\pm$0.0609  & \textbf{0.2715$\pm$0.0406}  & 0.2785$\pm$0.0684  & \textbf{0.2636$\pm$0.0483}  \\
& 3000 $\rightarrow$ 300 & 0.3547$\pm$0.0252  & \textbf{0.3479$\pm$0.0230}  & 0.3453$\pm$0.0096  & \textbf{0.3339$\pm$0.0061}  & 0.3444$\pm$0.0137  & \textbf{0.3372$\pm$0.0159}  & 0.3564$\pm$0.0237  & \textbf{0.3427$\pm$0.0161}  \\
& 3000 $\rightarrow$ 1000 & 0.4906$\pm$0.0069  & \textbf{0.4434$\pm$0.0050}  & 0.4919$\pm$0.0100  & \textbf{0.4727$\pm$0.0152}  & 0.4652$\pm$0.0062  & \textbf{0.4408$\pm$0.0193}  & 0.3528$\pm$0.0030  & \textbf{0.3501$\pm$0.0030}  \\

\midrule
  
\multirow{4}{*}{USHCN}
  & 50 $\rightarrow$ 3 & 0.4057$\pm$0.0164  & \textbf{0.4018$\pm$0.0159} & 0.4140$\pm$0.0301  & \textbf{0.4097$\pm$0.0291}  & 0.4576$\pm$0.0093  & \textbf{0.4548$\pm$0.0080}  & 0.4623$\pm$0.0081  & \textbf{0.4579$\pm$0.0093} \\
  & 100 $\rightarrow$ 3 & 0.4651$\pm$0.0095  & \textbf{0.4569$\pm$0.0108}  & 0.4561$\pm$0.0117  & \textbf{0.4539$\pm$0.0116}  & 0.2872$\pm$0.0609  & \textbf{0.2715$\pm$0.0406}  & 0.2785$\pm$0.0684  & \textbf{0.2636$\pm$0.0483} \\
  & 150 $\rightarrow$ 3 & 0.3547$\pm$0.0252  & \textbf{0.3479$\pm$0.0230}  & 0.3453$\pm$0.0096  & \textbf{0.3339$\pm$0.0061}  & 0.3444$\pm$0.0137  &  \textbf{0.3372$\pm$0.0159} & 0.3564$\pm$0.0237  & \textbf{0.3427$\pm$0.0161} \\
  & 150 $\rightarrow$ 50 & 0.4906$\pm$0.0069  & \textbf{0.4434$\pm$0.0050}  & 0.4919$\pm$0.0100  & \textbf{0.4727$\pm$0.0152}  & 0.4652$\pm$0.0062  & \textbf{0.4408$\pm$0.0193} & 0.3528$\pm$0.0030  & \textbf{0.3501$\pm$0.0030} \\

\bottomrule
\end{tabular}
}
\end{table*}

\section{Additional Experiments}
\label{sec:additional-experiments}

\subsection{Additional Main Results}
\label{sec:addtional-main-results}

We report additional forecasting results measured by the MAE metric in Table \ref{tab:main_results_mae}. \undercal improves performance across almost all models and datasets. On the Human Activity dataset, the MAE of SeFT decreases by 16.20\%, and that of Reformer drops by 21.51\%, mirroring the substantial reductions seen in MSE. Similarly, on USHCN, MIMIC, and PhysioNet, MAE improvements align with the corresponding MSE gains as well. These results further confirm that \undercal provides robust, model-agnostic enhancements for online irregular multivariate time series forecasting.

\subsection{Additional Ablation Study}
\label{sec:additional-ablation-study}

We further report ablation results of \undercal on additional source forecasters, including mTAN \cite{shukla2021multi}, Warpformer \cite{zhang2023warpformer}, HyperIMTS \cite{li2025hyperimts}, and GraFITi \cite{yalavarthi2024grafiti}, in Table \ref{tab:ablation_mtan}, \ref{tab:ablation_warpformer}, \ref{tab:ablation_hyperimts}, and \ref{tab:ablation_grafiti}. The results consistently show performance drops in most cases, confirming that all components jointly contribute to effective online adaptation across diverse source models.

\begin{table}
\centering
\caption{Ablation study of \undercal for mTAN on four datasets under five seeds evaluated using MSE.}
\label{tab:ablation_mtan}
\resizebox{\linewidth}{!}{
\begin{tabular}{lcccc}
\toprule
Method Variant & MIMIC & PhysioNet & Human Activity & USHCN \\
\midrule
w/o GDC (Single Expert, Joint) & 1.0363 & 0.4218 & 0.3529 & 0.4939 \\
w/o GDC (Single Expert, Reliable) & 0.9951 & 0.4225 & 0.3438 & 0.4890 \\
w/o GDC (Single Expert, Unreliable) & 1.0630 & 0.4274 & 0.4016 & 0.5129 \\
\midrule
w/o ARM (Random Triggering) & 0.9706 & 0.4264 & 0.3466 & 0.5085 \\
w/o ARM (Random Allocating) & 0.8955 & 0.4205 & 0.3489 & 0.4861 \\
\midrule
w/o All (Single Expert, Joint) & 1.0310 & 0.4217 & 0.3495 & 0.5012 \\
\midrule
\undercal & \textbf{0.8809} & \textbf{0.4193} & \textbf{0.3354} & \textbf{0.4792} \\
\bottomrule
\end{tabular}
}
\end{table}

\begin{table}
\centering
\caption{Ablation study of \undercal for Warpformer on four datasets under five seeds evaluated using MSE.}
\label{tab:ablation_warpformer}
\resizebox{\linewidth}{!}{
\begin{tabular}{lcccc}
\toprule
Method Variant & MIMIC & PhysioNet & Human Activity & USHCN \\
\midrule
w/o GDC (Single Expert, Joint) & 0.4692 & 0.3032 & 0.3310 & 0.4311 \\
w/o GDC (Single Expert, Reliable) & 0.4717 & 0.3033 & 0.3424 & 0.4299 \\
w/o GDC (Single Expert, Unreliable) & 0.4745 & 0.3038 & 0.3431 & 0.4354 \\
\midrule
w/o ARM (Random Triggering) & 0.4712 & 0.3034 & 0.3797 & 0.4323 \\
w/o ARM (Random Allocating) & 0.4723 & 0.3029 & 0.2976 & 0.4310 \\
\midrule
w/o All (Single Expert, Joint) & 0.4689 & 0.3031 & 0.3225 & 0.4304 \\
\midrule
\undercal & \textbf{0.4683} & \textbf{0.3028} & \textbf{0.3081} & \textbf{0.4295} \\
\bottomrule
\end{tabular}
}
\end{table}

\begin{table}
\centering
\caption{Ablation study of \undercal for HyperIMTS on four datasets under five seeds evaluated using MSE.}
\label{tab:ablation_hyperimts}
\resizebox{\linewidth}{!}{
\begin{tabular}{lcccc}
\toprule
Method Variant & MIMIC & PhysioNet & Human Activity & USHCN \\
\midrule
w/o GDC (Single Expert, Joint) & 0.5488 & 0.2983 & 0.0816 & 0.3973 \\
w/o GDC (Single Expert, Reliable) & 0.5484 & 0.2984 & 0.0815 & 0.3992 \\
w/o GDC (Single Expert, Unreliable) & 0.5544 & 0.2987 & 0.0839 & 0.4046 \\
\midrule
w/o ARM (Random Triggering) & 0.5552 & 0.2985 & 0.0817 & 0.4059 \\
w/o ARM (Random Allocating) & 0.5495 & 0.2982 & 0.0818 & 0.3985 \\
\midrule
w/o All (Single Expert, Joint) & \textbf{0.5456} & 0.2984 & 0.0816 & \textbf{0.3925} \\
\midrule
\undercal & 0.5457 & \textbf{0.2980} & \textbf{0.0813} & 0.3973 \\
\bottomrule
\end{tabular}
}
\end{table}

\begin{table}
\centering
\caption{Ablation study of \undercal for GraFITi on four datasets under five seeds evaluated using MSE.}
\label{tab:ablation_grafiti}
\resizebox{\linewidth}{!}{
\begin{tabular}{lcccc}
\toprule
Method Variant & MIMIC & PhysioNet & Human Activity & USHCN \\
\midrule
w/o GDC (Single Expert, Joint) & 0.4513 & 0.2992 & 0.0790 & 0.5520 \\
w/o GDC (Single Expert, Reliable) & 0.4520 & 0.2993 & 0.0790 & 0.5811 \\
w/o GDC (Single Expert, Unreliable) & 0.4589 & 0.2996 & 0.0796 & 0.5627 \\
\midrule
w/o ARM (Random Triggering) & 0.4524 & 0.2994 & 0.0793 & 0.5877 \\
w/o ARM (Random Allocating) & 0.4527 & 0.2992 & 0.0792 & 0.5537 \\
\midrule
w/o All (Single Expert, Joint) & 0.4513 & 0.2992 & \textbf{0.0788} & 0.5726 \\
\midrule
\undercal & \textbf{0.4482} & \textbf{0.2989} & 0.0790 & \textbf{0.5546} \\
\bottomrule
\end{tabular}
}
\end{table}

\subsection{Additional Lookback Length and Horizon Study}
\label{sec:additional-lookback}

We report additional MAE results for varying lookback lengths and forecast horizons in Table \ref{tab:vary_window_results_mae}. \undercal consistently reduces MAE across most of lookback lengths, forecast horizons. Meanwhile, improvements are observed on most models and datasets, demonstrating that our method effectively leverages historical information and maintains stability under extended forecast horizons.

\subsection{Uncertainty Estimator Robustness Analysis}
\label{sec:UE-robustness}

To simulate radical online distribution shifts and evaluate the robustness of \undercal under inaccurate uncertainty estimation, we inject Gaussian noise with variance $\sigma^2 = 0.5$ into UE outputs. Results in Table \ref{tab:ue-robustness} show that, despite strong perturbations, performance re-stabilizes within 1--2 batches while maintaining an average MSE improvement of about 0.5\%, demonstrating strong self-correction and stable online adaptation capabilities.

\begin{table}
\caption{MSE change under inaccurate uncertainty estimation on Human Activity.}
\small
\begin{tabular*}{\linewidth}{@{\extracolsep{\fill}}cccccc}
\toprule
Batch & 1 & 2 & 3 & 4 & 5 \\
\midrule
MSE Change & -0.12\% & +0.01\% & +0.30\% & +0.33\% & +0.60\% \\
\bottomrule
\end{tabular*}
\label{tab:ue-robustness}
\end{table}

\subsection{Discussion on Gated Distribution Calibrator}

To analyze the performance drop of \textit{w/o GDC (Single Expert, Joint}, we conduct a gradient conflict analysis. We observe that \textit{w/o GDC (Single Expert, Reliable/Unreliable)} performs worse than the joint variant, suggesting that challenging samples remain informative rather than acting purely as noise. Moreover, approximately 45\% of training batches exhibit gradient conflicts between reliable and unreliable groups. In contrast, \undercal achieves better performance, indicating that dedicated experts effectively mitigate optimization interference.

\subsection{Trigger Threshold Sensitivity Analysis}

To evaluate the trade-off between update frequency and forecasting performance, we vary the trigger threshold $\tau_{\text{trig}}$ and report the results in Table \ref{tab:trigger-threshold}. We observe that MSE reaches a plateau when $\tau_{\text{trig}} \leq 0.75$, while update frequency continues increasing. Therefore, our chosen setting reduces update frequency by approximately 15\% without noticeable performance loss.

\begin{table}
\centering
\caption{Trade-off between trigger threshold, update frequency, and performance on Human Activity for tPatchGNN.}
\small
\begin{tabular}{ccc}
\toprule
$\tau_{\text{trig}}$ & MSE & Frequency (\%) \\
\midrule
0.25 & 0.1269 & 96.47 \\
0.50 & 0.1266 & 92.71 \\
0.75 & 0.1258 & 85.41 \\
1.00 & 0.1280 & 79.47 \\
1.50 & 0.1278 & 76.47 \\
2.00 & 0.1300 & 58.82 \\
3.00 & 0.1376 & 21.18 \\
\bottomrule
\end{tabular}
\label{tab:trigger-threshold}
\end{table}

\section{Baseline Details}
\label{sec:baseline-details}

We briefly introduce the baseline methods used in our experiments. For both regular and irregular multivariate time series forecasting models, we adopt the implementations from the PyOmniTS benchmark\footnote{\url{https://github.com/Ladbaby/PyOmniTS}}. More details can be found in HyperIMTS~\cite{li2025hyperimts} and ReIMTS~\cite{li2026reimts}.
For online learning and test-time adaptation methods:
\begin{itemize}    
    \item \textbf{OneNet}~\cite{wen2023onenet}: An online deep learning model that integrates cross-variable and cross-time modeling to capture dynamic dependencies.
    \item \textbf{FSNet}~\cite{pham2022learning}: An online forecasting framework with a dual-stream mechanism that leverages both immediate and delayed feedback for adaptive learning.
    \item \textbf{Detect-and-Adapt} (D3A)~\cite{zhang2024addressing}: A concept drift adaptation method that explicitly detects distribution shifts and updates model parameters accordingly.
    \item \textbf{TAFAS}~\cite{kim2025battling}: A test-time adaptation framework that utilizes partially observed ground truths to proactively calibrate predictions.
\end{itemize}

\end{document}